\documentclass[fleqn,10pt]{wlscirep}
\usepackage[utf8]{inputenc}
\usepackage[T1]{fontenc}

\usepackage{color}

\usepackage{booktabs}
\usepackage{multirow}
\newcommand{\datacell}[2]{\begin{tabular}{@{}c@{}}$#1$\\#2\end{tabular}}
\newcommand{\datacellt}[2]{\begin{tabular}{@{}c@{}}#1\\#2\end{tabular}}

\usepackage{graphicx}
\usepackage{subcaption}

\usepackage{microtype}

\title{Value-Based Large Language Model Agent Simulation for Mutual Evaluation of Trust and Interpersonal Closeness}

\author[1,*]{Yuki Sakamoto}
\author[1]{Takahisa Uchida}
\author[1]{Hiroshi Ishiguro}
\affil[1]{The University of Osaka, Graduate School of Engineering Science, Toyonaka 5608531, Japan}

\affil[*]{sakamoto.yuki@irl.sys.es.osaka-u.ac.jp}

\begin{abstract}
Large language models (LLMs) have emerged as powerful tools for simulating complex social phenomena using human-like agents with specific traits.
In human societies, value similarity is important for building trust and close relationships; however, it remains unexplored whether this principle holds true in artificial societies comprising LLM agents. 
Therefore, this study investigates the influence of value similarity on relationship-building among LLM agents through two experiments. 
First, in a preliminary experiment, we evaluated the controllability of values in LLMs to identify the most effective model and prompt design for controlling the values. 
Subsequently, in the main experiment, we generated pairs of LLM agents imbued with specific values and analyzed their mutual evaluations of trust and interpersonal closeness following a dialogue. 
The experiments were conducted in English and Japanese to investigate language dependence. 
The results confirmed that pairs of agents with higher value similarity exhibited greater mutual trust and interpersonal closeness. 
Our findings demonstrate that the LLM agent simulation serves as a valid testbed for social science theories, contributes to elucidating the mechanisms by which values influence relationship building, and provides a foundation for inspiring new theories and insights into the social sciences.
\end{abstract}

\begin{document}

\flushbottom
\maketitle
%
%
\thispagestyle{empty}

\section*{Introduction}

Recently, the rapid advancement of large language models (LLMs) has opened new possibilities for social simulations using artificial agents. 
Pretrained on vast amounts of data, LLMs can act as specific personas through natural language instructions (prompts), owing to their in-context learning capabilities \cite{brown2020language,jiang2023personallm}. 
This characteristic has facilitated their application in multiagent simulations, which model complex social phenomena by efficiently generating diverse agents and observing their interactions \cite{park2023generative}. 
Studies have begun to assign personalities and roles to agents to analyze emergent phenomena and collective task-solving abilities \cite{qian2023communicative,piao2025agentsociety}. 
Thus, previous studies have focused primarily on the behavioral characteristics and social roles of agents, contributing to the replication of social phenomena and elucidation of cooperative behavior.

With this trend, the introduction of values as a fundamental driver of agent behavior has garnered attention in the pursuit of human-like principles of action. 
Values are defined as abstract, enduring beliefs that guide the choices and evaluations of an individual across situations \cite{schwartz1992universals,schwartz2012overview}, thereby serving as a central concept in understanding human behavior. 
In the context of LLM agent research, studies based on Schwartz's theory of basic human values have demonstrated the potential to control agent values via prompts, showing that agents imbued with specific values tend to exhibit the corresponding behaviors \cite{kovavc2023large,kovavc2024stick}.

Furthermore, values play a decisive role not only in individual decision-making but also in social interactions \cite{hitlin2004values}. 
Value similarity is crucial in fostering intimate interpersonal relationships and promoting trust \cite{murstein1970stimulus,siegrist2000salient}. 
However, most existing studies on value-based LLM agents have been limited to verifying the controllability of values or analyzing their impact on task performance. 
The question of how values, particularly their similarity, influence relationship formation (e.g., trust) in social interactions between agents remains largely unexplored.

Therefore, this study simulates the influence of value similarity on the formation of relationships through the interactions between value-based LLM agents. 
The primary objective of this study is to verify whether the principle that higher value similarity enhances mutual attraction and leads to the formation of positive relationships holds true in an artificial society comprising LLM agents. 
Furthermore, the relative importance and structure of values \cite{schwartz1992universals,markus2014culture}, along with the trust formation processes \cite{yamagishi1994trust}, can differ between cultures. 
Accordingly, we focus on Japan—a culture that emphasizes interpersonal relationships and group harmony more than several Western societies \cite{markus2014culture}—and explore the cultural influences mediated by language by performing simulations in both English and Japanese environments. 
Through these investigations, the present study contributes to the development of more realistic multiagent social simulations in which diverse values coexist. 
In addition, this study aims not only to reproduce existing findings from social sciences but also to provide new insights into the mechanisms of relationship formation from our simulation results back to the social sciences, thereby inspiring new theoretical perspectives.

\begin{figure}[ht]
\centering
\includegraphics[width=0.8\linewidth]{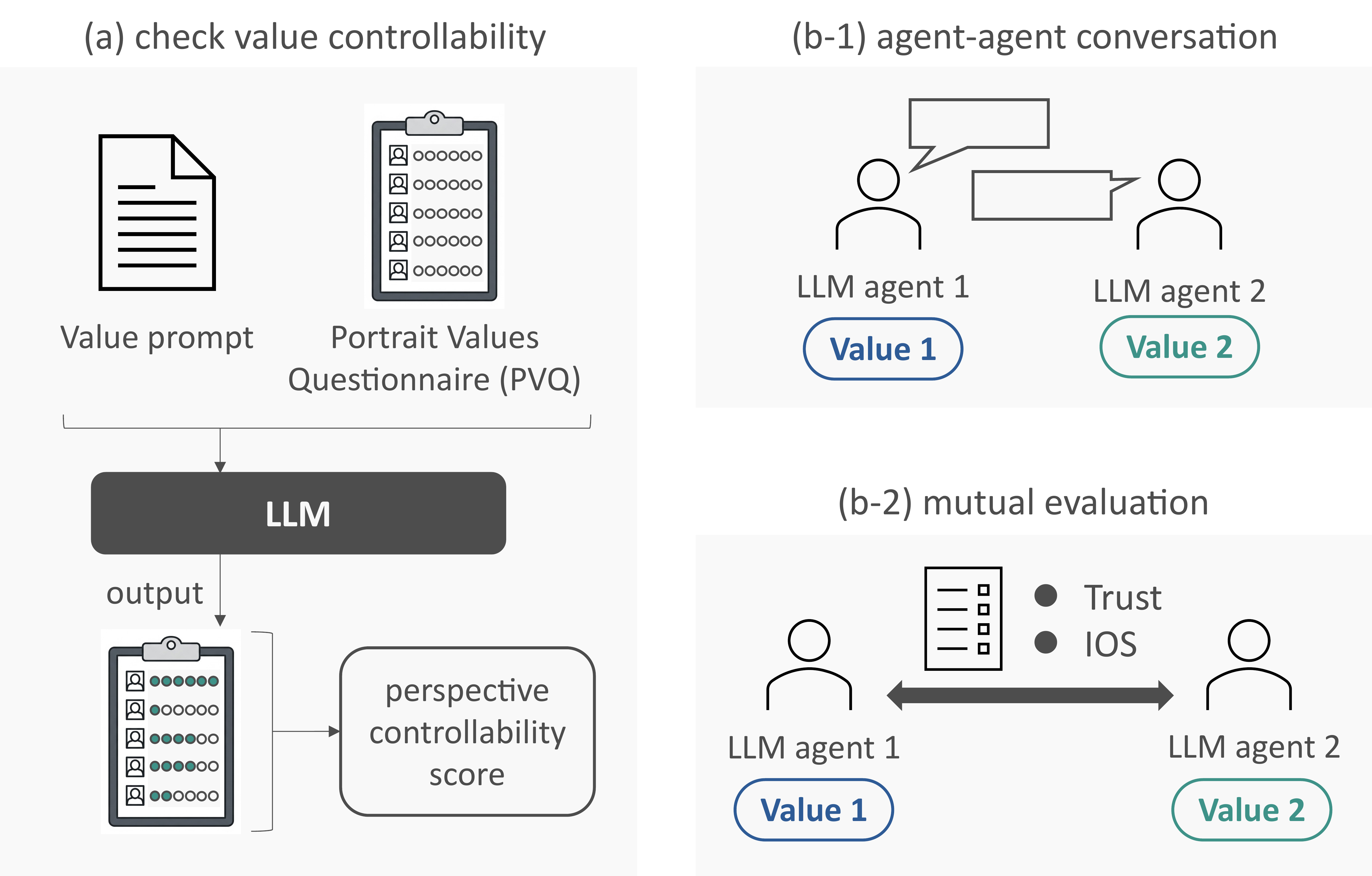}
\caption{Overall experimental workflow. (a) The preliminary experiment assesses the value controllability of LLMs, whereas (b) the main experiment investigates the effect of value similarity on mutual evaluation.}
\label{fig:overview}
\end{figure}

A two-stage experiment was designed to achieve these objectives. 
The overall flow of the experiment is shown in Figure \ref{fig:overview}. 
In a preliminary experiment, we quantitatively assessed the controllability of values through prompts for multiple LLMs, following the methodology proposed by a previous study \cite{kovavc2023large}. 
In the main experiment, we investigated the influence of value similarity on interagent relationships. 
We generated pairs of agents with specific values using an LLM, for which value control was confirmed in the preliminary experiment. 
After engaging in dialogue, they evaluated each other using a questionnaire designed to measure mutual trust and interpersonal closeness.

The contributions of this study are as follows:
\begin{itemize}
    \item We identified optimal prompt designs for LLM value control by comprehensively validating the controllability of values across different languages and prompt designs. In particular, demonstrating the feasibility of controlling LLM values in a Japanese context provides crucial data for expanding computational social science studies.
    \item We quantitatively demonstrated that the value similarity principle, which is crucial for trust and relationship building, emerges in an artificial society constructed with LLM agents. Furthermore, by comparing English and Japanese environments, we elucidated both the universal and language-dependent aspects of the influence of values on relationship formation.
    \item We demonstrated that LLM agent simulations can serve as a viable experimental platform for testing existing social science theories. Beyond this, we proposed the potential for a bidirectional research loop between artificial intelligence and the social sciences, where new patterns discovered through simulations can provide insights into new hypotheses about human social behavior to the social sciences.
\end{itemize}

The remainder of this paper is organized as follows. 
First, we present the results of a preliminary experiment on the controllability of LLM values and a main experiment on the relationship between value similarity and relationship building. 
Next, we provide a discussion of the experimental results. 
Finally, we describe the experimental details.

\section*{Results}

\subsection*{Preliminary Experiment}

This preliminary experiment evaluated the value controllability of LLMs via prompting. 
We quantified the LLM responses to instructions to adopt specific values using a perspective controllability score, which indicates the extent to which the model expresses the value or personal trait corresponding to the prompted perspective more strongly than other traits \cite{kovavc2023large}. 
Controllability was assessed across each value dimension (the ten basic and four higher-order values in Schwartz's value theory), with parameters including prompt design (system or user prompt, second-person or third-person prompt, inclusion or exclusion of value definitions) and language (English and Japanese).
The system prompt refers to the instruction primarily used to set the model's overall behavior and role, whereas the user prompt is the input provided by the user during the interaction. These are standard functionalities provided in the APIs of large language models (LLMs).
In the second-person prompt, the model is addressed directly using ``you.'' In the third-person prompt, the prompt starts with the phrase ``The following are answers from a person ...''.
The experimental procedure is described in detail in the Methods section.

The measured perspective controllability scores are presented in Table \ref{tab:controllability_basic} for the ten basic values and in Table \ref{tab:controllability_higher} for the four higher-order values.
For the ten basic values, the highest controllability was observed under the combination of ``Gemini, system prompt, second-person, with definition,'' in both English and Japanese.
For the four higher-order values, the highest controllability was observed under ``Gemini, user prompt, second-person, with definition'' in English, and under ``Gemini, system prompt, second-person, without definition'' in Japanese.

Following prior work \cite{kovavc2023large}, we conducted Welch's t-tests with Bonferroni correction to compare controllability across models.
Using the prompt design that yielded the highest controllability for each LLM, we compared Gemini—the model with the highest controllability—against Claude and GPT. 
For the ten basic values, Gemini exhibited significantly higher controllability than both Claude (\textit{t}(62.76) = 136.19, \textit{p} < .001) and GPT (\textit{t}(57.81) = 106.05, \textit{p} < .001) in English, as well as significantly higher controllability than Claude (\textit{t}(97.31) = 52.85, \textit{p} < .001) and GPT (\textit{t}(78.52) = 40.62, \textit{p} < .001) in Japanese. 
For the four higher-order values, Gemini again outperformed Claude (\textit{t}(68.32) = 49.19, \textit{p} < .001) and GPT (\textit{t}(78.84) = 69.37, \textit{p} < .001) in English, and also showed significantly higher controllability than Claude (\textit{t}(61.14) = 76.06, \textit{p} < .001) and GPT (\textit{t}(63.64) = 44.31, \textit{p} < .001) in Japanese.
Furthermore, to compare controllability between languages, we conducted Welch's t-tests with Bonferroni correction using the prompt design and LLM that achieved the highest controllability in each language. 
There was a significant difference between English and Japanese for both the ten basic values, \emph{t}(51.69) = 53.90, \emph{p} < .001, and the four higher-order values, \emph{t}(97.94) = 87.13, \emph{p} < .001.

\begin{table}[ht]
\centering
\begin{tabular}{@{}ll l cc cc cc@{}}
\toprule
& & & \multicolumn{2}{c}{\textbf{Gemini}} & \multicolumn{2}{c}{\textbf{Claude}} & \multicolumn{2}{c}{\textbf{GPT}} \\
\cmidrule(lr){4-5} \cmidrule(lr){6-7} \cmidrule(lr){8-9}
\textbf{Prompt} & \textbf{Person} & \textbf{Definition} & English & Japanese & English & Japanese & English & Japanese \\
\midrule
\multirow[c]{4}{*}[-2.5em]{System}%
  & \multirow[c]{2}{*}[-0.85em]{2} & Yes & \datacell{\textbf{0.8216}\pm}{0.0027} & \datacell{\textbf{0.6961}\pm}{0.0161} & --- & --- & \datacell{0.5931\pm}{0.0048} & \datacell{0.4729\pm}{0.0146} \\
  \cmidrule(lr){3-9}
  &                                 & No  & \datacell{0.7315\pm}{0.0089} & \datacell{0.6237\pm}{0.0037} & --- & --- & \datacell{0.5772\pm}{0.0078} & --- \\
  \cmidrule(lr){2-9}
  & \multirow[c]{2}{*}[-0.85em]{3} & Yes & \datacell{0.7678\pm}{0.0029} & \datacell{0.4784\pm}{0.0168} & \datacell{0.6251\pm}{0.0063} & --- & \datacell{0.6119\pm}{0.0039} & \datacell{0.4982\pm}{0.0195} \\
  \cmidrule(lr){3-9}
  &                                 & No  & \datacell{0.7123\pm}{0.0183} & \datacell{0.3442\pm}{0.0180} & \datacell{\textbf{0.6752}\pm}{0.0070} & \datacell{\textbf{0.5166}\pm}{0.0175} & \datacell{0.6014\pm}{0.0076} & \datacell{\textbf{0.5097}\pm}{0.0278} \\
\midrule
\multirow[c]{4}{*}[-2.5em]{User}%
  & \multirow[c]{2}{*}[-0.85em]{2} & Yes & \datacell{0.8142\pm}{0.0049} & \datacell{0.6959\pm}{0.0120} & --- & --- & \datacell{0.6552\pm}{0.0051} & --- \\
  \cmidrule(lr){3-9}
  &                                 & No  & \datacell{0.7692\pm}{0.0074} & \datacell{0.6269\pm}{0.0159} & --- & --- & \datacell{0.6485\pm}{0.0077} & --- \\
  \cmidrule(lr){2-9}
  & \multirow[c]{2}{*}[-0.85em]{3} & Yes & --- & --- & --- & --- & \datacell{0.6731\pm}{0.0059} & --- \\
  \cmidrule(lr){3-9}
  &                                 & No  & --- & --- & --- & --- & \datacell{\textbf{0.6816}\pm}{0.0088} & --- \\
\bottomrule
\end{tabular}
\caption{Perspective controllability scores in the ten basic values. The table reports the mean and standard deviation of the scores obtained from 50 runs. The highest value in each column is shown in bold.}
\label{tab:controllability_basic}
\end{table}

\begin{table}[ht]
\centering
\begin{tabular}{@{}ll l cc cc cc@{}}
\toprule
& & & \multicolumn{2}{c}{\textbf{Gemini}} & \multicolumn{2}{c}{\textbf{Claude}} & \multicolumn{2}{c}{\textbf{GPT}} \\
\cmidrule(lr){4-5} \cmidrule(lr){6-7} \cmidrule(lr){8-9}
\textbf{Prompt} & \textbf{Person} & \textbf{Definition} & English & Japanese & English & Japanese & English & Japanese \\
\midrule
\multirow[c]{4}{*}[-2.5em]{System}%
  & \multirow[c]{2}{*}[-0.85em]{2} & Yes & \datacell{0.8766\pm}{0.0035} & \datacell{0.7796\pm}{0.0076} & --- & --- & \datacell{0.6836\pm}{0.0086} & \datacell{0.6071\pm}{0.0281} \\
  \cmidrule(lr){3-9}
  &                                 & No  & \datacell{0.8768\pm}{0.0090} & \datacell{\textbf{0.7824}\pm}{0.0058} & --- & --- & \datacell{0.6823\pm}{0.0089} & --- \\
  \cmidrule(lr){2-9}
  & \multirow[c]{2}{*}[-0.85em]{3} & Yes & \datacell{0.8432\pm}{0.0048} & \datacell{0.5949\pm}{0.0103} & --- & --- & \datacell{0.6937\pm}{0.0080} & \datacell{0.6643\pm}{0.0223} \\
  \cmidrule(lr){3-9}
  &                                 & No  & \datacell{0.8361\pm}{0.0062} & \datacell{0.5733\pm}{0.0122} & \datacell{\textbf{0.7870}\pm}{0.0125} & \datacell{\textbf{0.5929}\pm}{0.0164} & \datacell{0.7038\pm}{0.0090} & \datacell{\textbf{0.6811}\pm}{0.0149} \\
\midrule
\multirow[c]{4}{*}[-2.5em]{User}%
  & \multirow[c]{2}{*}[-0.85em]{2} & Yes & \datacell{\textbf{0.8838}\pm}{0.0057} & \datacell{0.7647\pm}{0.0128} & --- & --- & \datacell{0.7483\pm}{0.0090} & \datacell{0.6782\pm}{0.0193} \\
  \cmidrule(lr){3-9}
  &                                 & No  & \datacell{0.8553\pm}{0.0079} & \datacell{0.7636\pm}{0.0125} & --- & --- & \datacell{0.7463\pm}{0.0094} & --- \\
  \cmidrule(lr){2-9}
  & \multirow[c]{2}{*}[-0.85em]{3} & Yes & --- & --- & --- & --- & \datacell{0.7618\pm}{0.0074} & --- \\
  \cmidrule(lr){3-9}
  &                                 & No  & --- & --- & --- & \datacell{0.5903\pm}{0.0189} & \datacell{\textbf{0.7719}\pm}{0.0098} & --- \\
\bottomrule
\end{tabular}
\caption{Perspective controllability scores in the 4 higher-order values. The table reports the mean and standard deviation of the scores obtained from 50 runs. The highest value in each column is shown in bold.}
\label{tab:controllability_higher}
\end{table}

\subsection*{Main Experiment}

We conducted a simulation experiment to investigate the effect of value similarity on the mutual evaluation between agents. 
In this experiment, we used two agents employing a combination of prompt design and an LLM, which demonstrated the highest value controllability in the preliminary experiment. 
The agents evaluated each other's trust \cite{nakayachi2014method} and interpersonal closeness (using the Inclusion of Other in the Self (IOS) scale \cite{aron1992inclusion}) after their dialogue. 
We set two topics for the dialogue: hobbies and collaborative decision-making regarding housing. 
Prompts, dialogue, and evaluations were performed in both English and Japanese.

\begin{figure*}[tb]
    \centering
    \begin{subfigure}[b]{0.45\linewidth}
        \centering
        \includegraphics[width=0.95\linewidth]{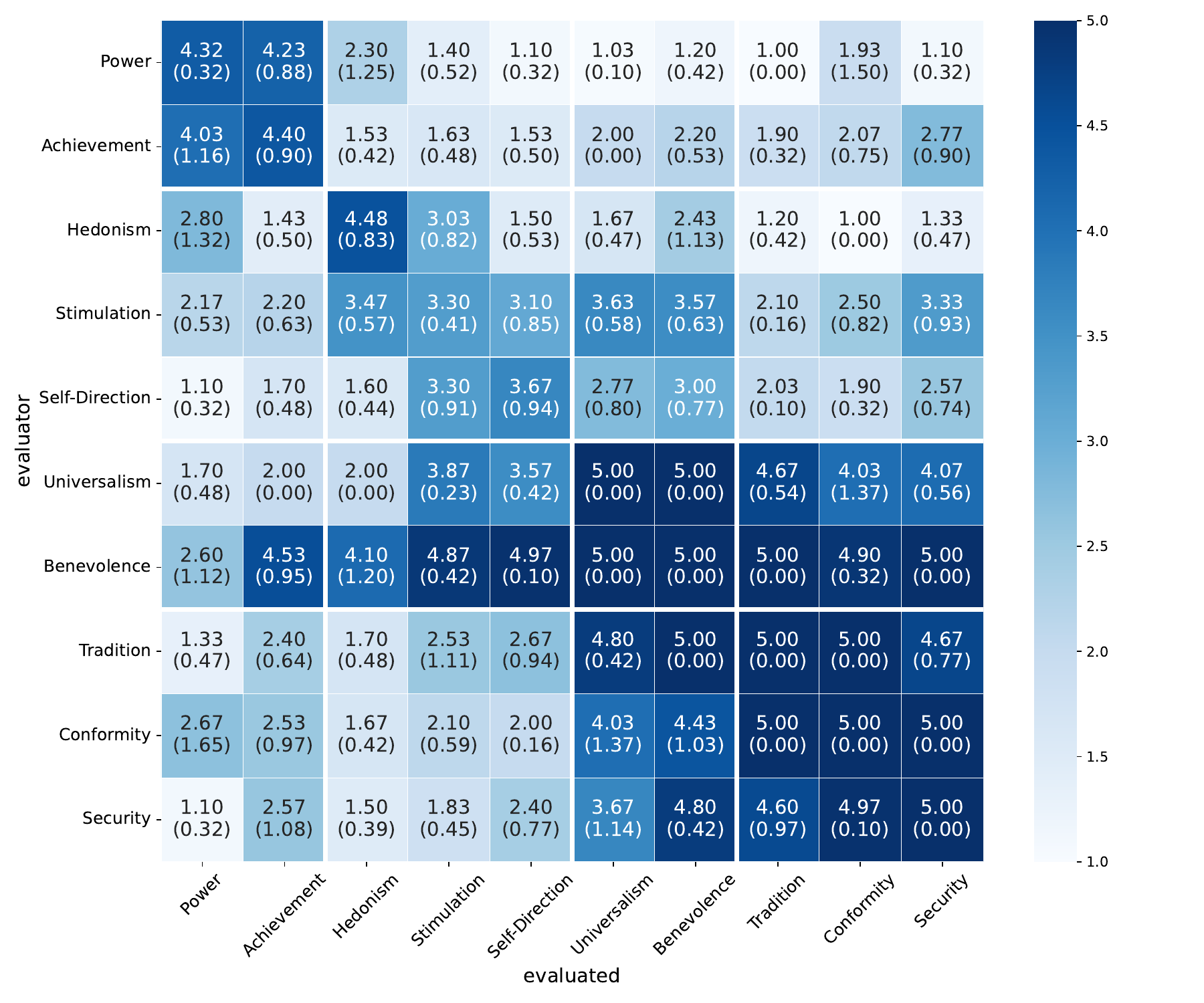}
        \caption{Trust, basic values (M = 3.151, SD = 1.516).}
        \label{fig:english_house_trust_basic}
    \end{subfigure}
    \hfill
    \begin{subfigure}[b]{0.45\linewidth}
        \centering
        \includegraphics[width=0.95\linewidth]{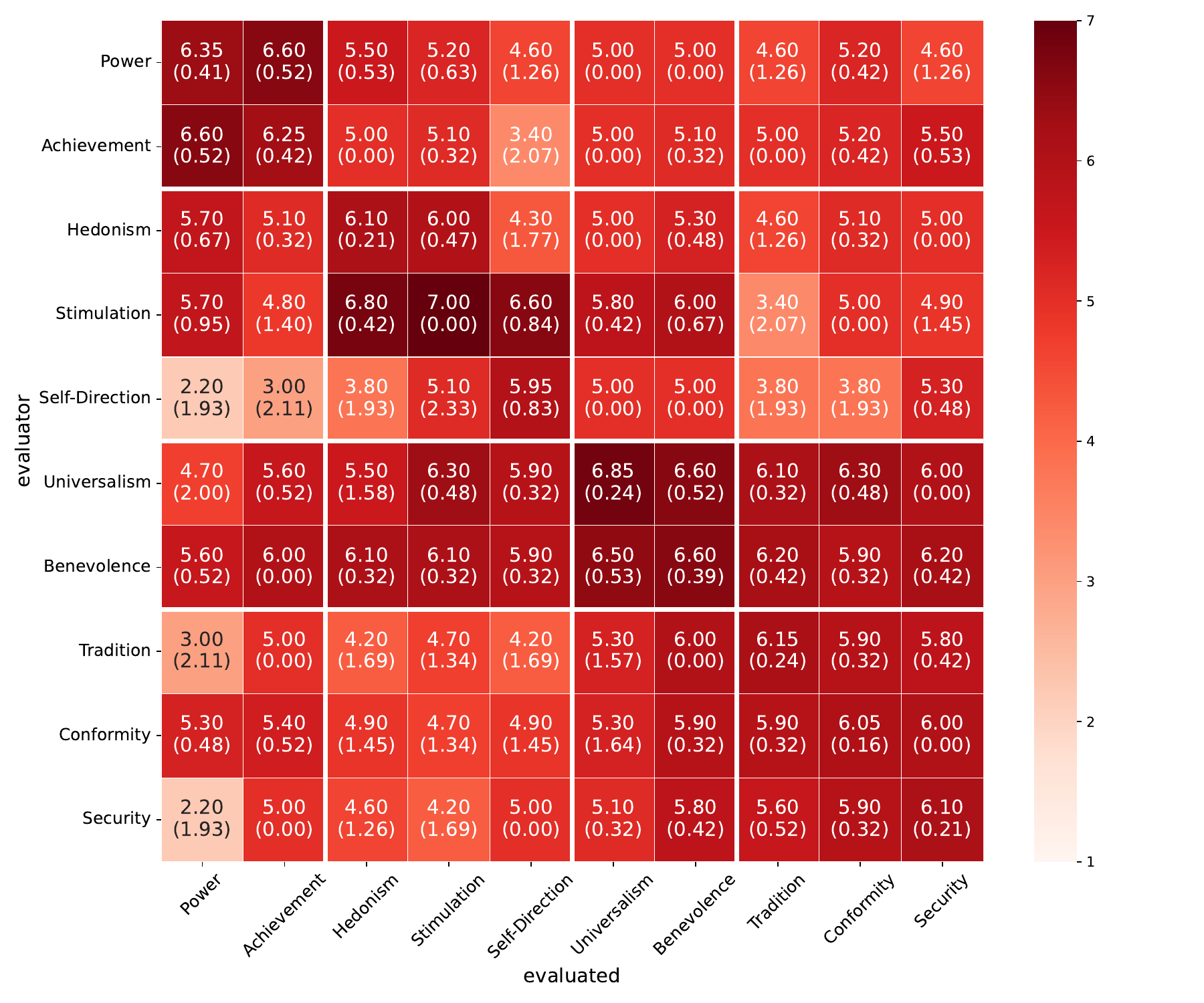}
        \caption{IOS, basic values (M = 5.404, SD = 1.310).}
        \label{fig:english_house_IOS_basic}
    \end{subfigure}

    \vspace{5mm}

    \begin{subfigure}[b]{0.45\linewidth}
        \centering
        \includegraphics[width=0.95\linewidth]{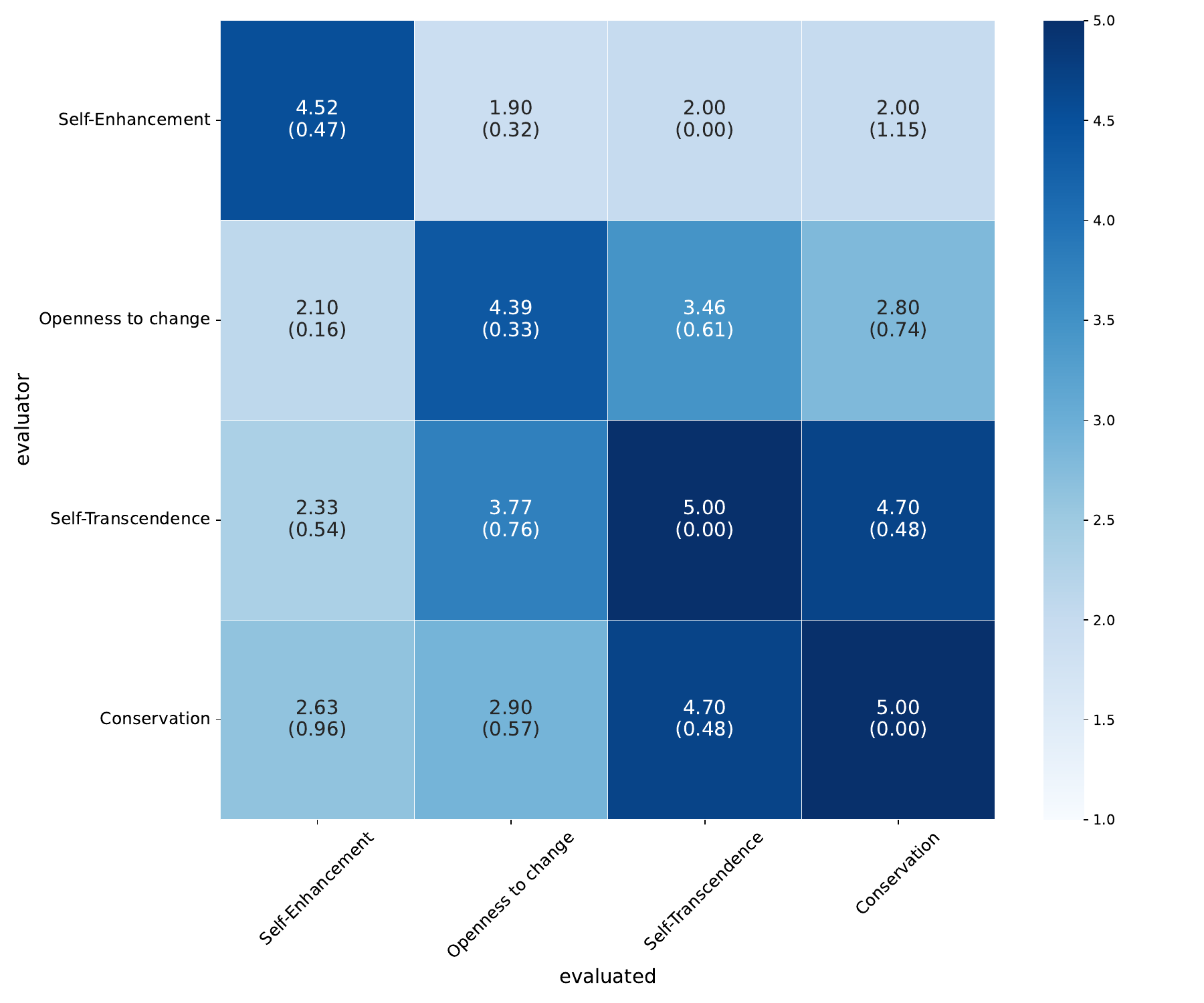}
        \caption{Trust, higher-order values (M = 3.655, SD = 1.277).}
        \label{fig:english_house_trust_higher}
    \end{subfigure}
    \hfill
    \begin{subfigure}[b]{0.45\linewidth}
        \centering
        \includegraphics[width=0.95\linewidth]{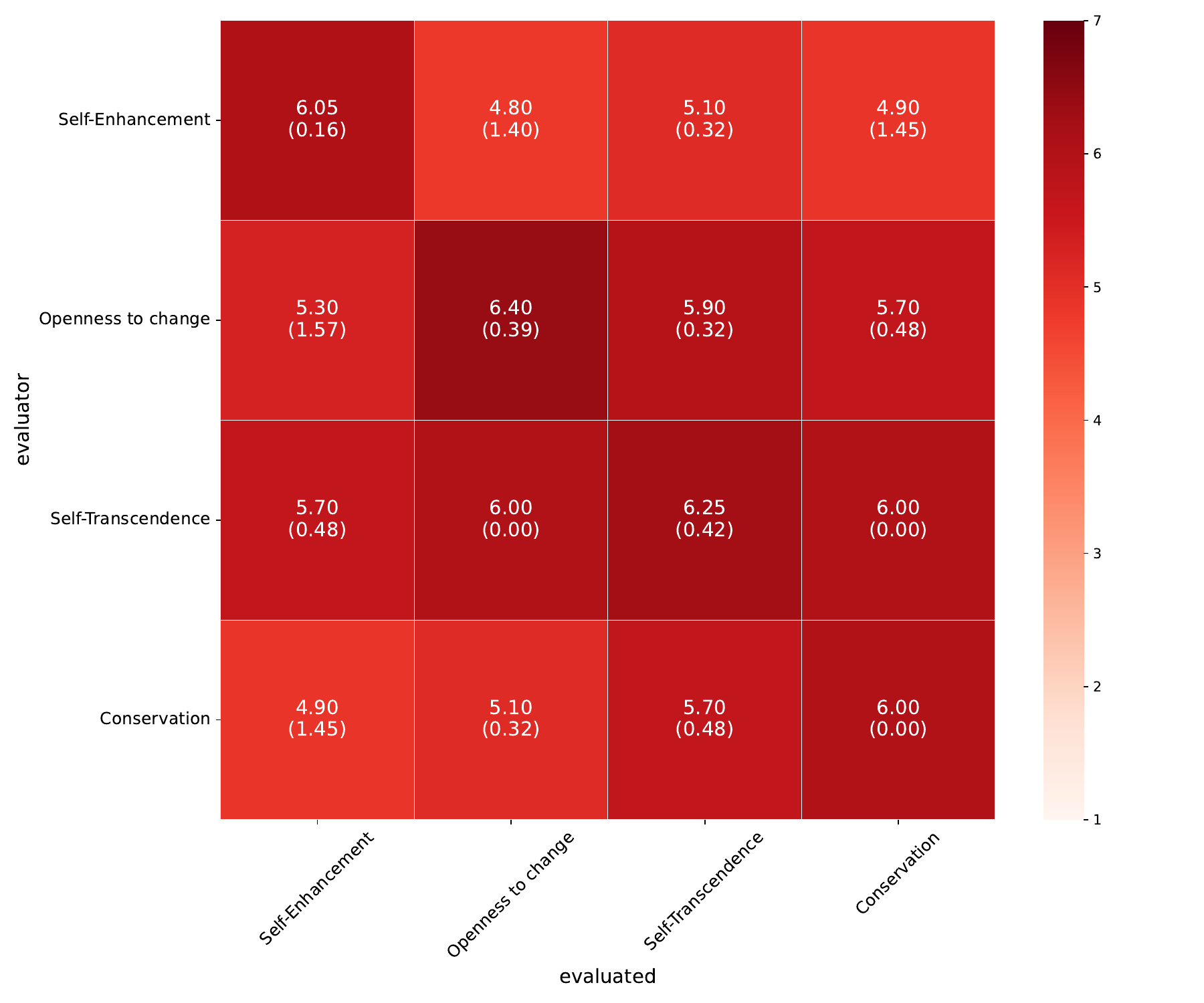}
        \caption{IOS, higher-order values (M = 5.725, SD = 0.862).}
        \label{fig:english_house_IOS_higher}
    \end{subfigure}
    
    \caption{Results of mutual evaluation (collaborative decision-making about housing, English). Each cell shows the mean score across 10 independent runs, with the standard deviation in parentheses. Rows represent the evaluator agent's value, and columns represent the evaluated agent's value.}
    \label{fig:english_house_matrix}
\end{figure*}

\begin{figure*}[tb]
    \centering
    \begin{subfigure}[b]{0.45\linewidth}
        \centering
        \includegraphics[width=0.95\linewidth]{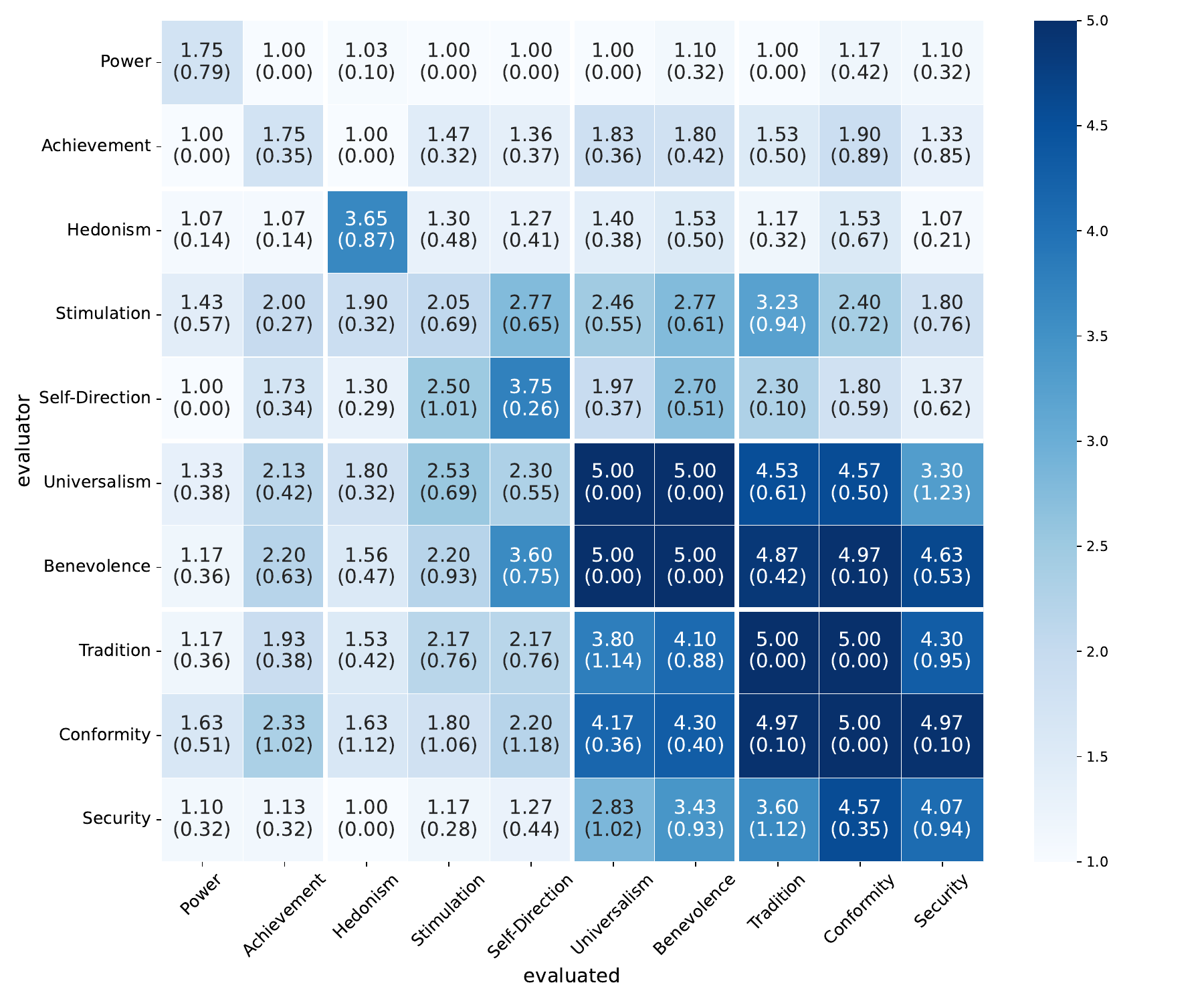}
        \caption{Trust, basic values (M = 2.522, SD = 1.497).}
        \label{fig:english_hobby_trust_basic}
    \end{subfigure}
    \hfill
    \begin{subfigure}[b]{0.45\linewidth}
        \centering
        \includegraphics[width=0.95\linewidth]{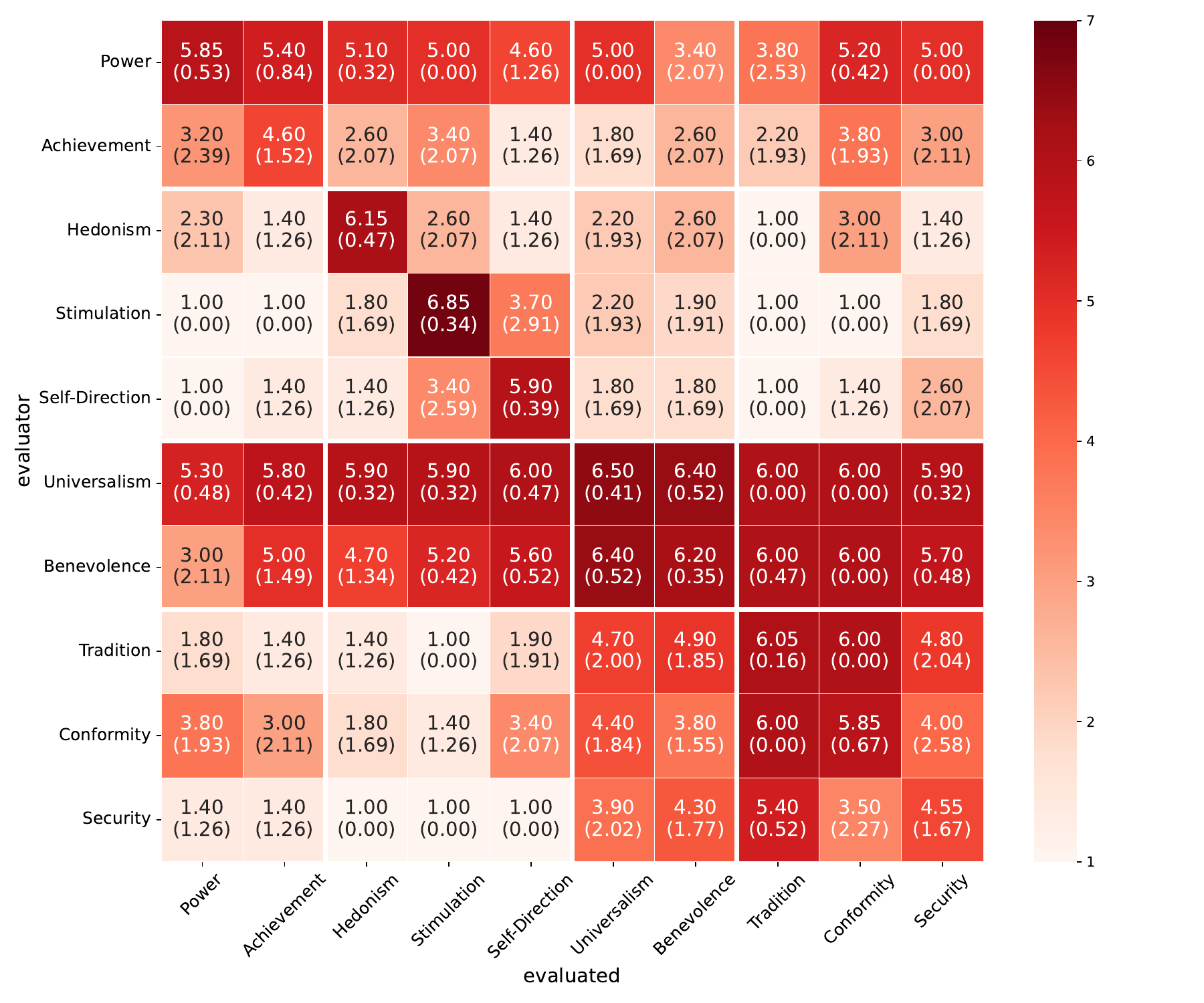}
        \caption{IOS, basic values (M = 3.780, SD = 2.323).}
        \label{fig:english_hobby_IOS_basic}
    \end{subfigure}

    \vspace{5mm}

    \begin{subfigure}[b]{0.45\linewidth}
        \centering
        \includegraphics[width=0.95\linewidth]{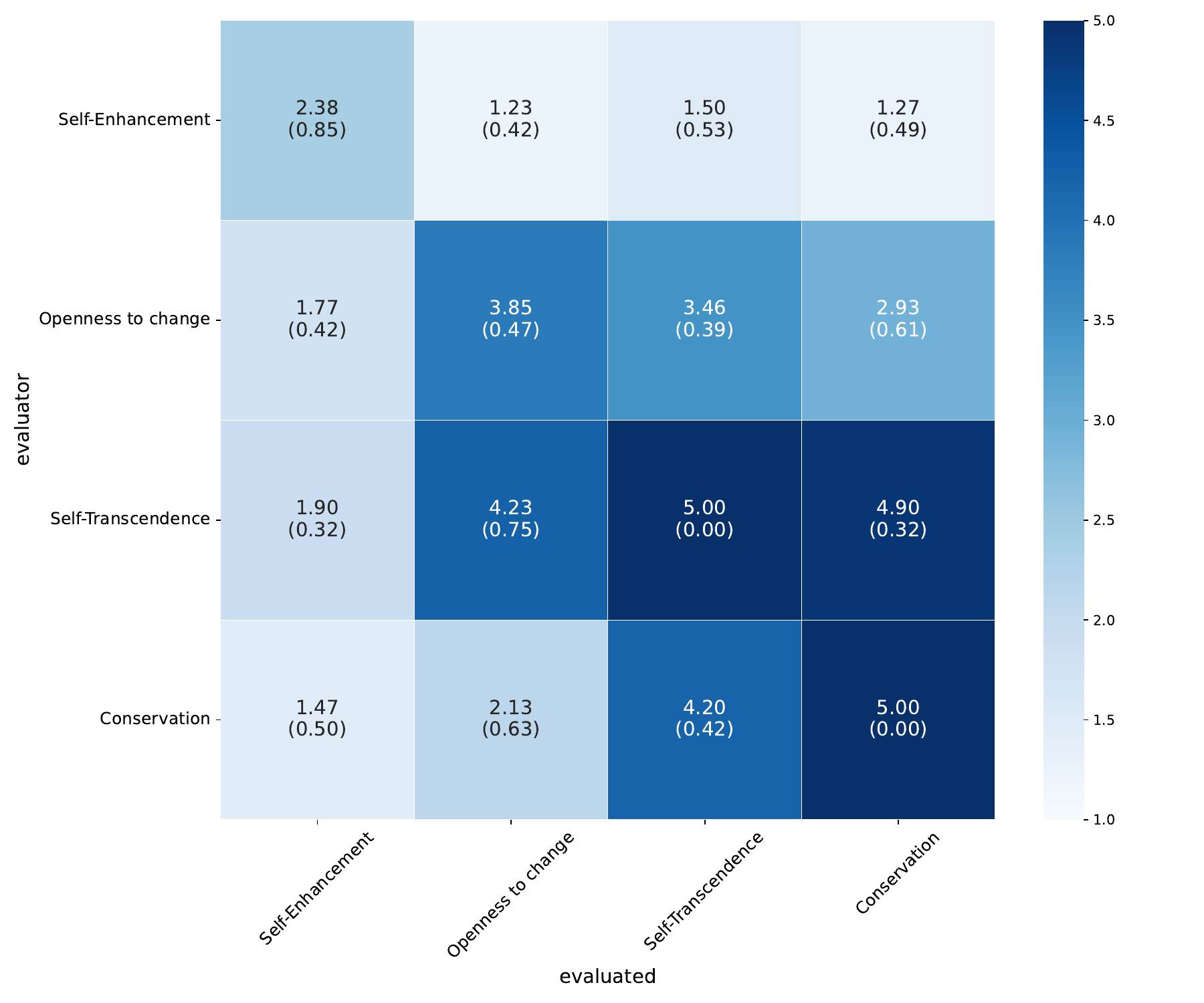}
        \caption{Trust, higher-order values (M = 3.174, SD = 1.471).}
        \label{fig:english_hobby_trust_higher}
    \end{subfigure}
    \hfill
    \begin{subfigure}[b]{0.45\linewidth}
        \centering
        \includegraphics[width=0.95\linewidth]{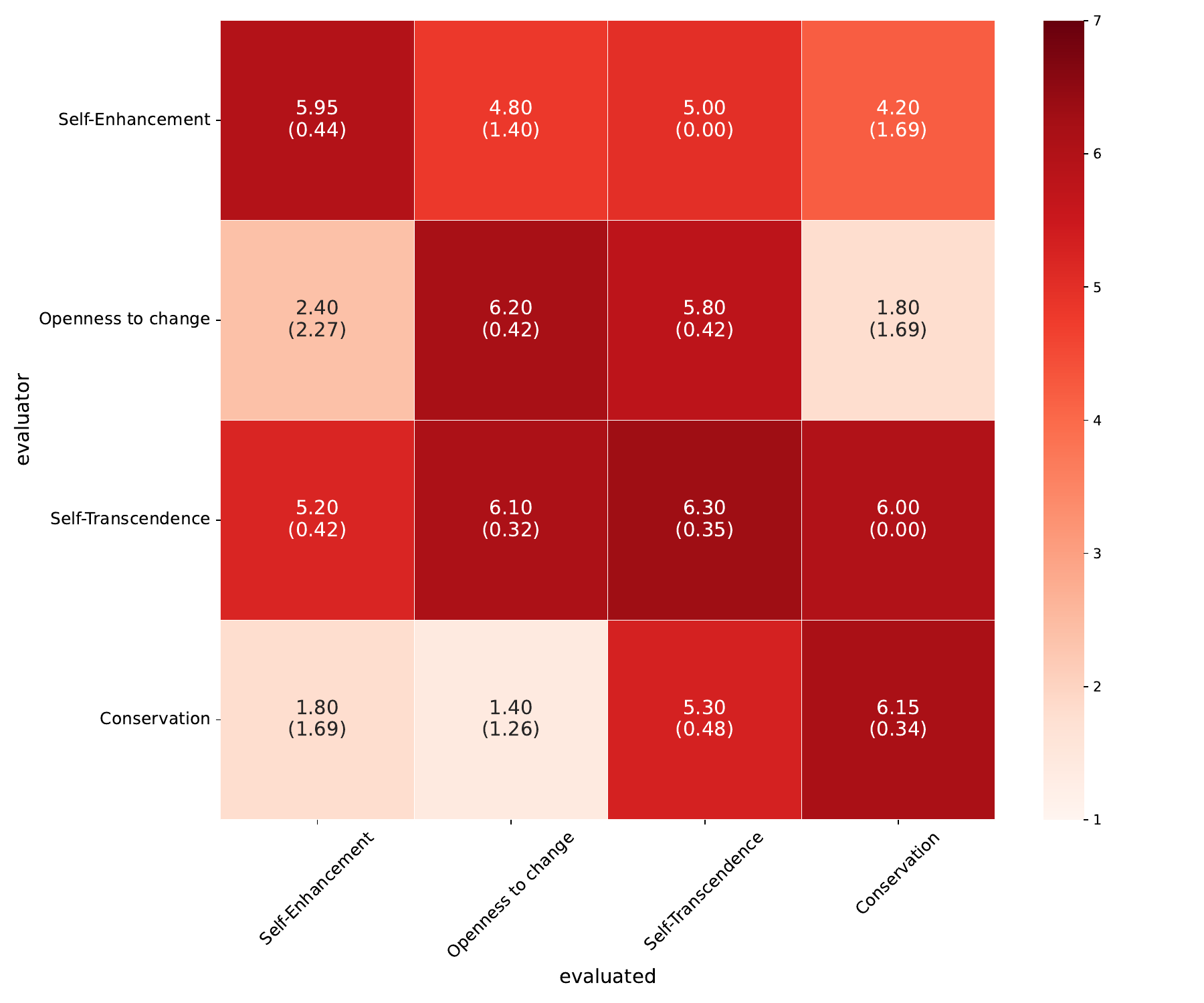}
        \caption{IOS, higher-order values (M = 4.950, SD = 1.904).}
        \label{fig:english_hobby_IOS_higher}
    \end{subfigure}
    
    \caption{Results of mutual evaluation (dialogue about hobbies, English). Each cell shows the mean score across 10 independent runs, with the standard deviation in parentheses. Rows represent the evaluator agent's value, and columns represent the evaluated agent's value.}
    \label{fig:english_hobby_matrix}
\end{figure*}

\begin{figure*}[tb]
    \centering
    \begin{subfigure}[b]{0.45\linewidth}
        \centering
        \includegraphics[width=0.95\linewidth]{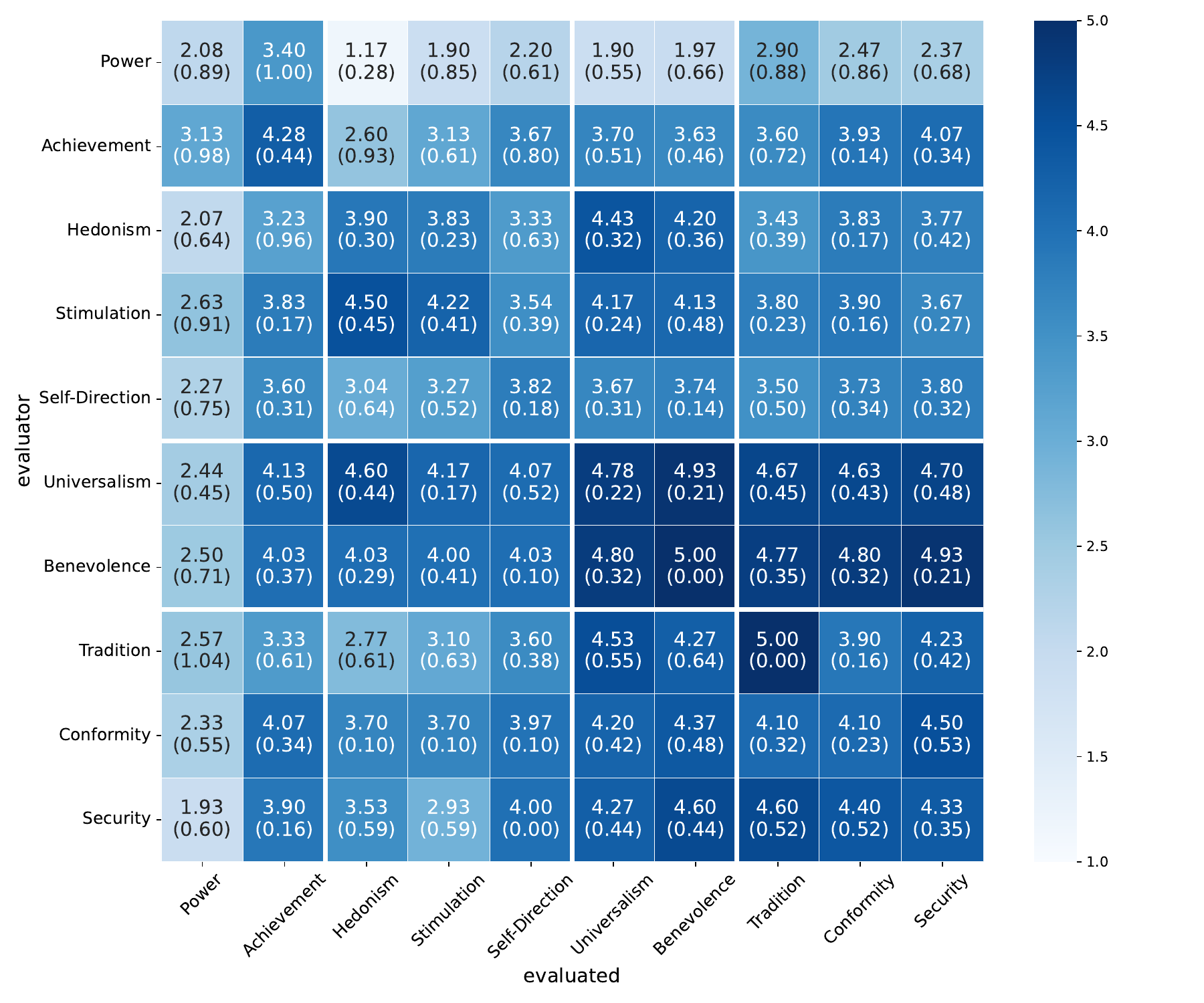}
        \caption{Trust, basic values (M = 3.721, SD = 0.971).}
        \label{fig:japanese_house_trust_basic}
    \end{subfigure}
    \hfill
    \begin{subfigure}[b]{0.45\linewidth}
        \centering
        \includegraphics[width=0.95\linewidth]{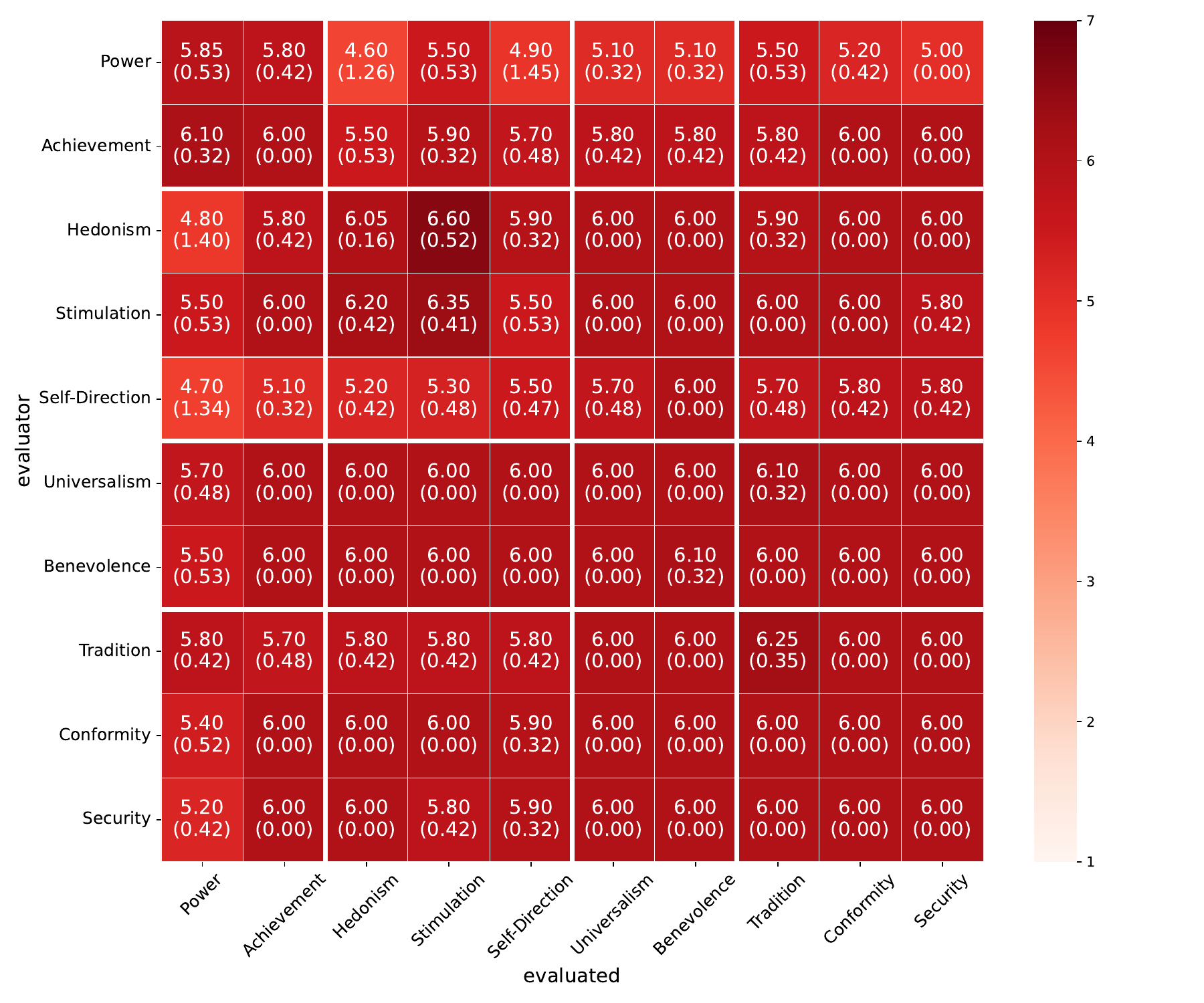}
        \caption{IOS, basic values (M = 5.829, SD = 0.519).}
        \label{fig:japanese_house_IOS_basic}
    \end{subfigure}

    \vspace{5mm}

    \begin{subfigure}[b]{0.45\linewidth}
        \centering
        \includegraphics[width=0.95\linewidth]{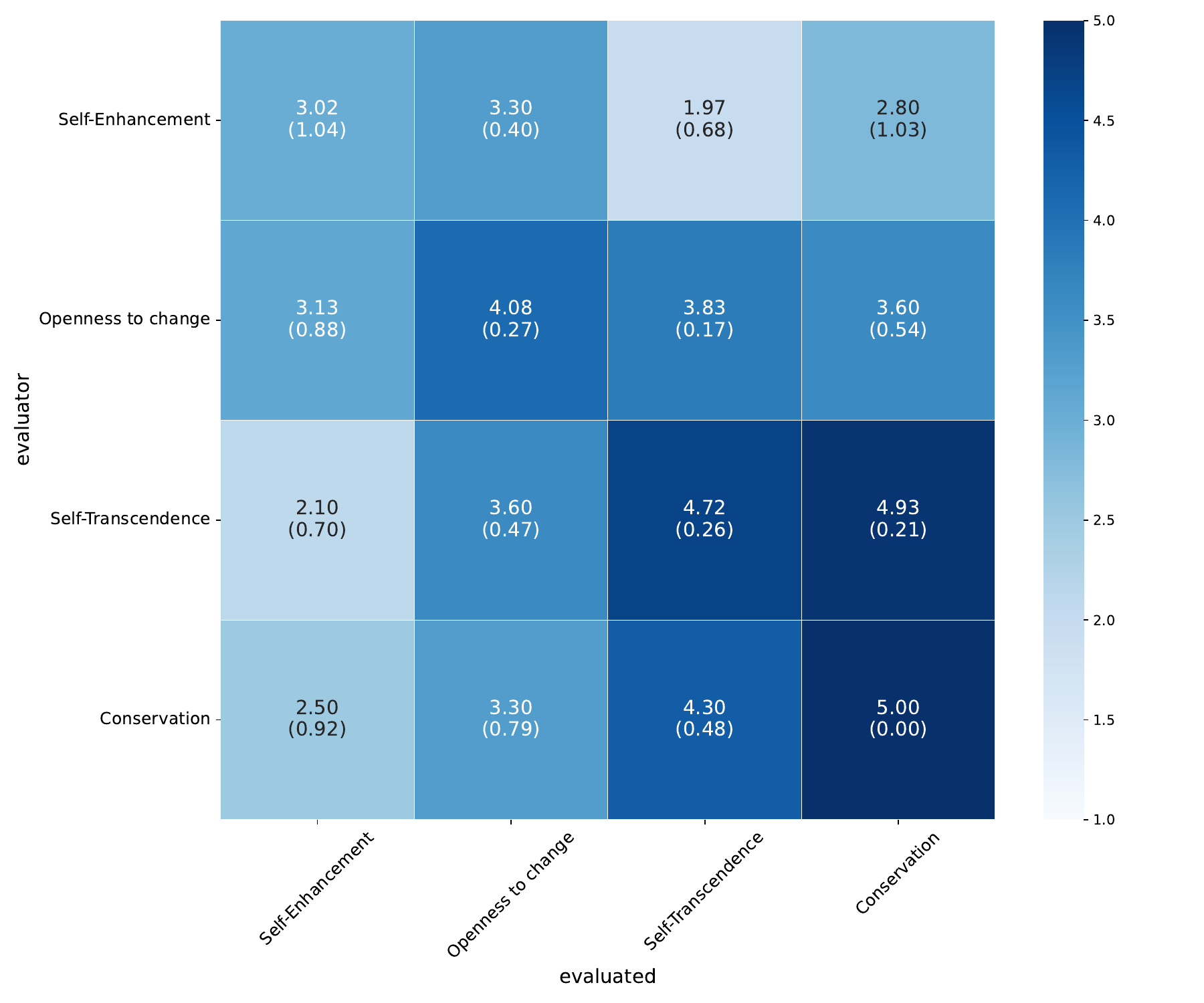}
        \caption{Trust, higher-order values (M = 3.651, SD = 1.108).}
        \label{fig:japanese_house_trust_higher}
    \end{subfigure}
    \hfill
    \begin{subfigure}[b]{0.45\linewidth}
        \centering
        \includegraphics[width=0.95\linewidth]{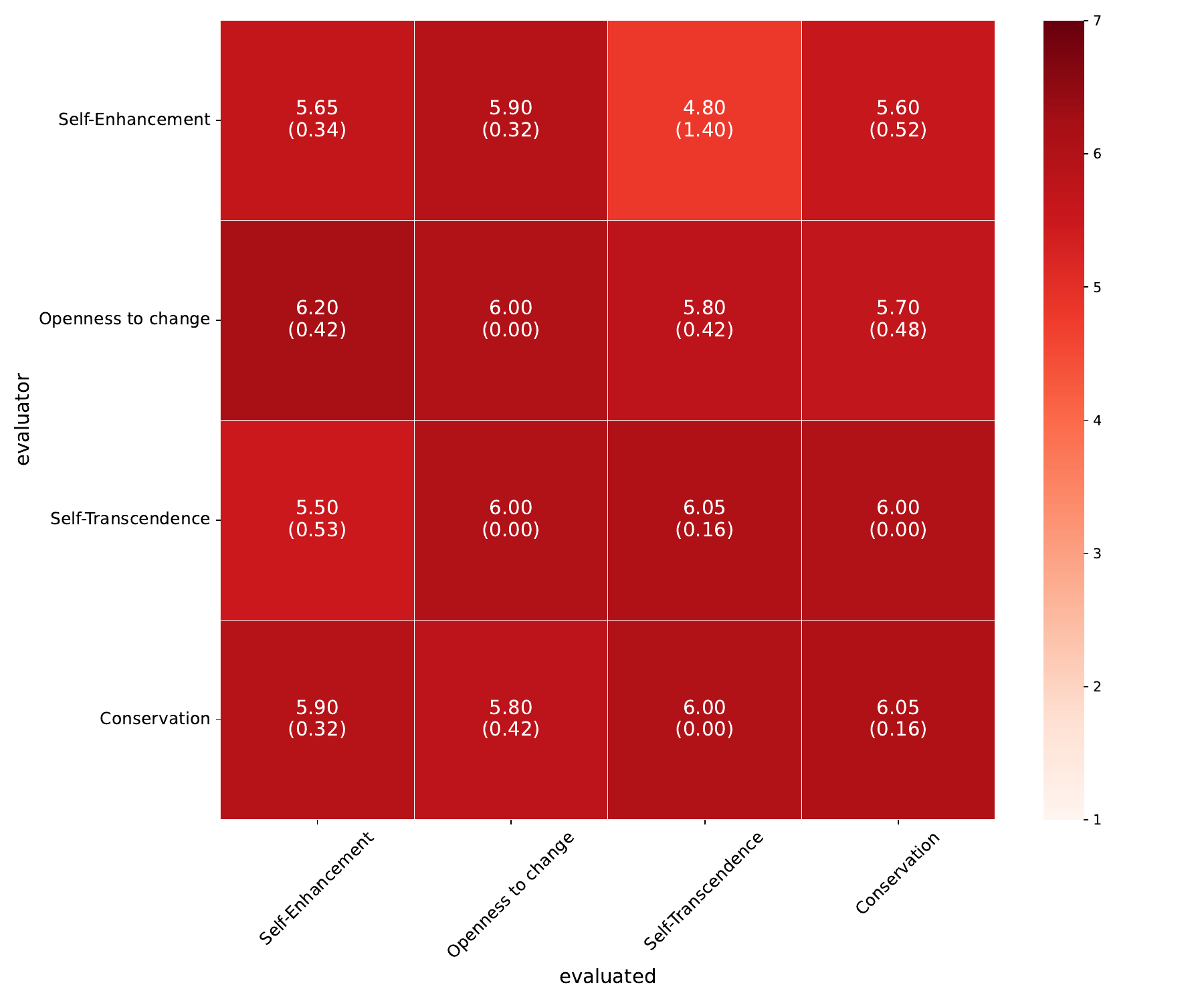}
        \caption{IOS, higher-order values (M = 5.835, SD = 0.538).}
        \label{fig:japanese_house_IOS_higher}
    \end{subfigure}
    
    \caption{Results of mutual evaluation (collaborative decision-making about housing, Japanese). Each cell shows the mean score across 10 independent runs, with the standard deviation in parentheses. Rows represent the evaluator agent's value, and columns represent the evaluated agent's value.}
    \label{fig:japanese_house_matrix}
\end{figure*}

\begin{figure*}[tb]
    \centering
    \begin{subfigure}[b]{0.45\linewidth}
        \centering
        \includegraphics[width=0.95\linewidth]{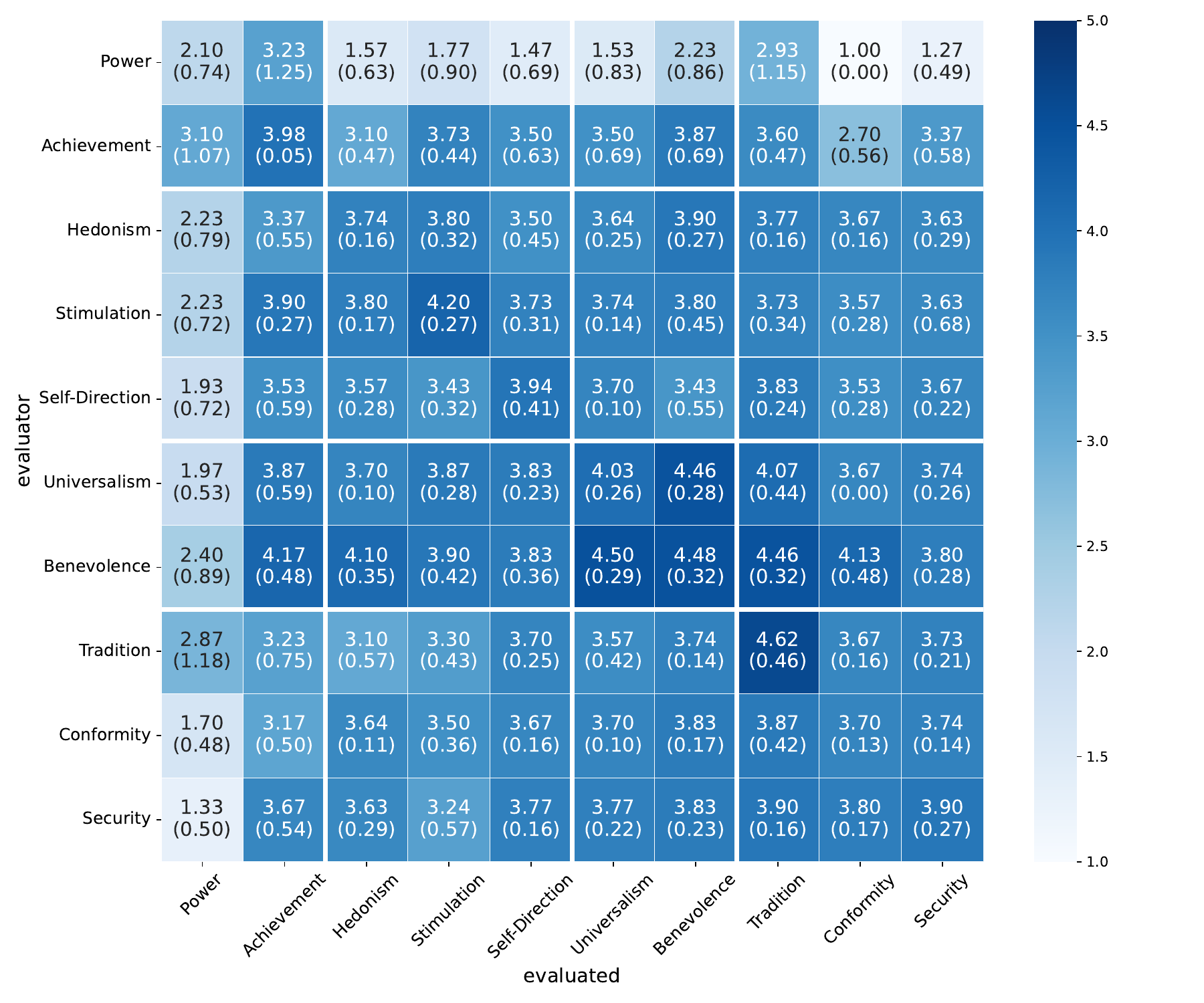}
        \caption{Trust, basic values (M = 3.457, SD = 0.900).}
        \label{fig:japanese_hobby_trust_basic}
    \end{subfigure}
    \hfill
    \begin{subfigure}[b]{0.45\linewidth}
        \centering
        \includegraphics[width=0.95\linewidth]{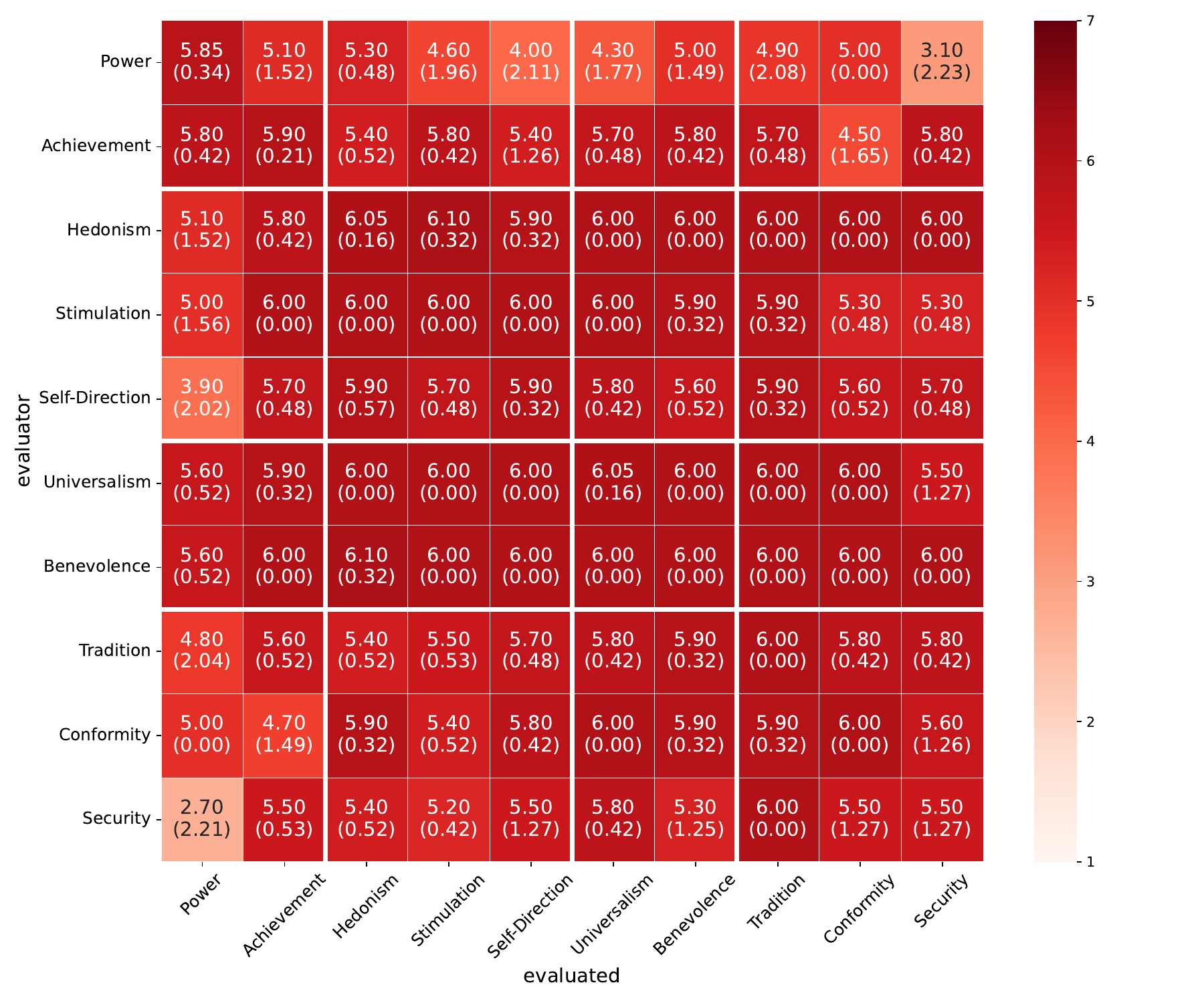}
        \caption{IOS, basic values (M = 5.617, SD = 0.956).}
        \label{fig:japanese_hobby_IOS_basic}
    \end{subfigure}

    \vspace{5mm}

    \begin{subfigure}[b]{0.45\linewidth}
        \centering
        \includegraphics[width=0.95\linewidth]{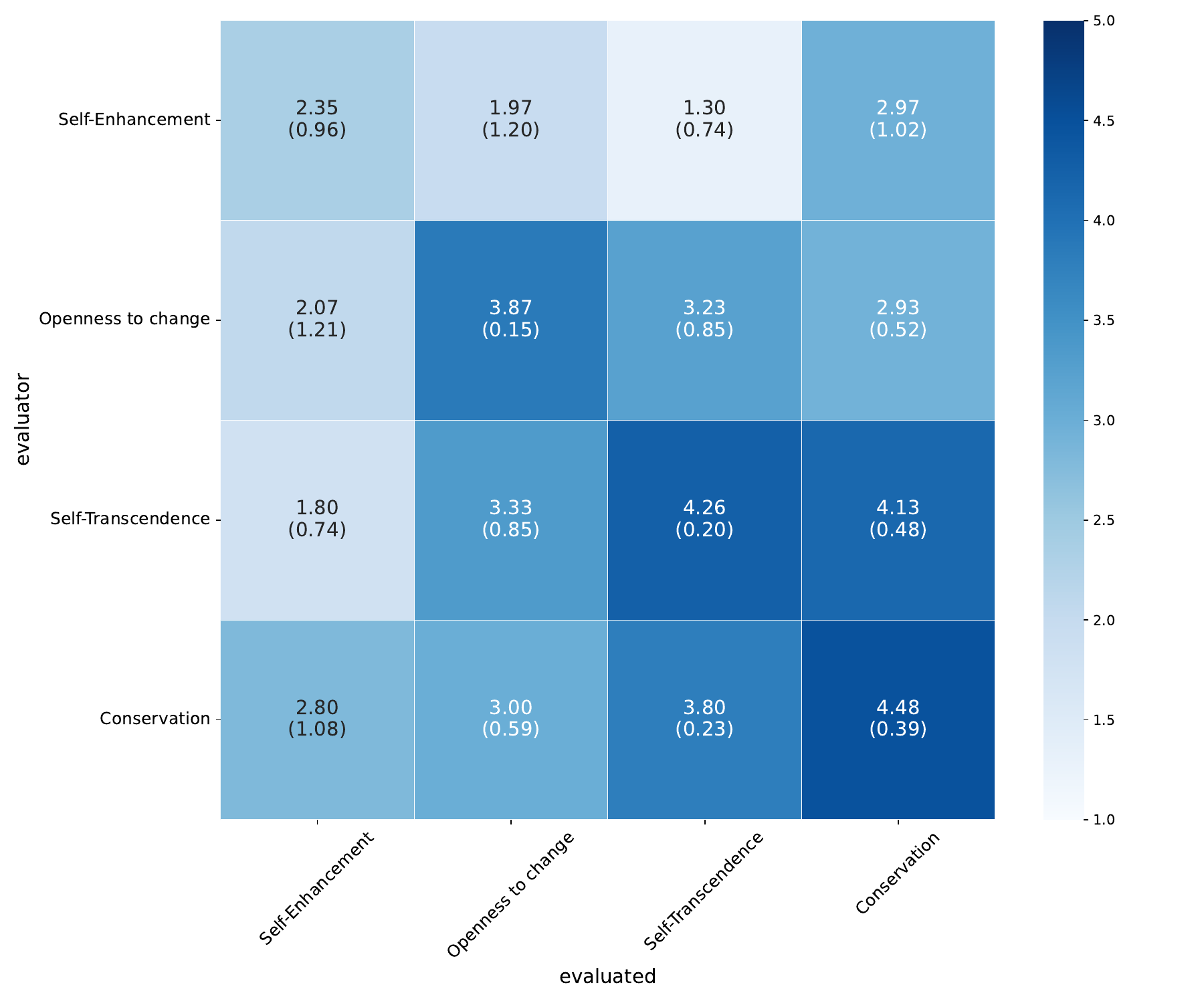}
        \caption{Trust, higher-order values (M = 3.164, SD = 1.187).}
        \label{fig:japanese_hobby_trust_higher}
    \end{subfigure}
    \hfill
    \begin{subfigure}[b]{0.45\linewidth}
        \centering
        \includegraphics[width=0.95\linewidth]{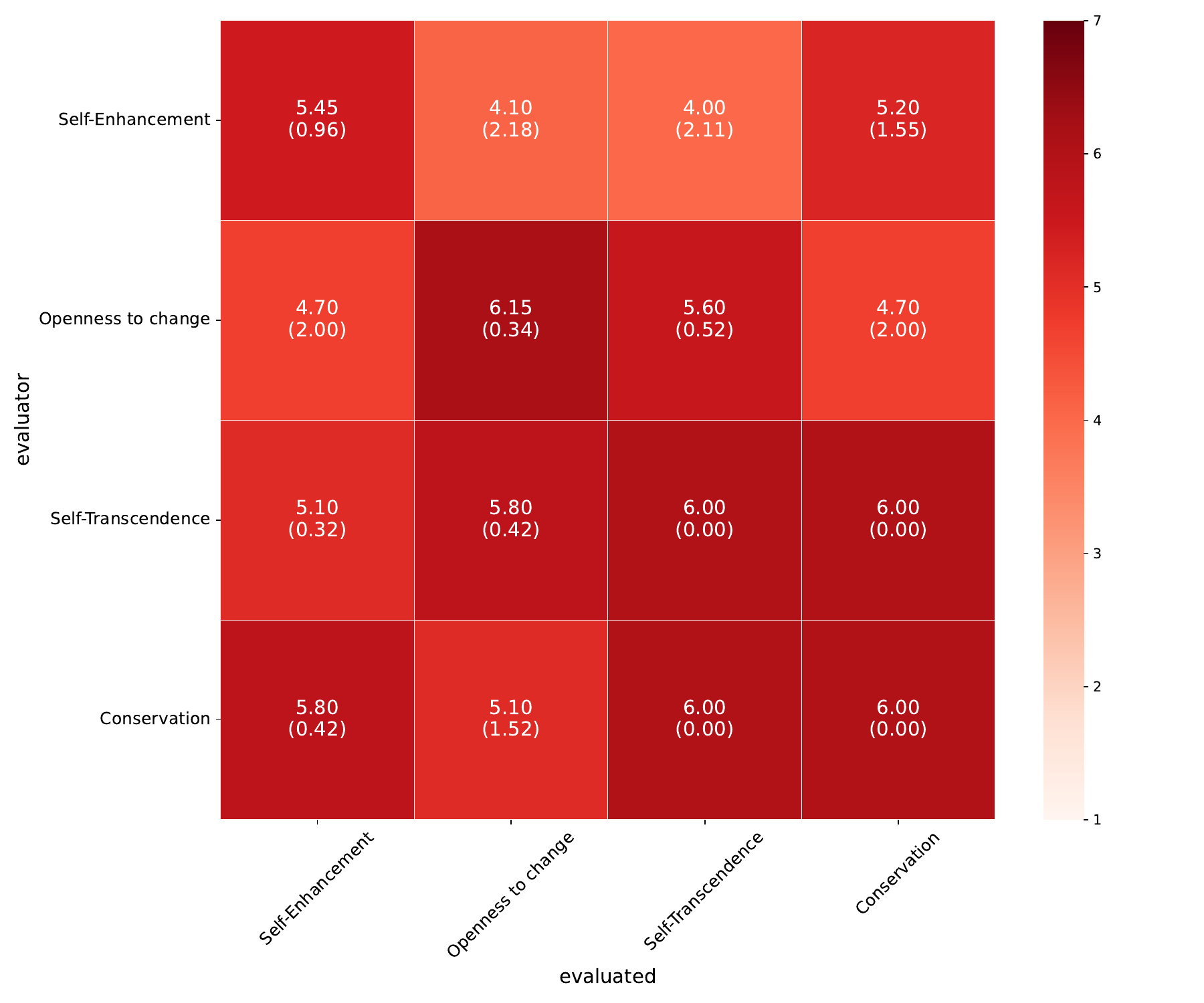}
        \caption{IOS, higher-order values (M = 5.465, SD = 1.260).}
        \label{fig:japanese_hobby_IOS_higher}
    \end{subfigure}
    
    \caption{Results of mutual evaluation (dialogue about hobbies, Japanese). Each cell shows the mean score across 10 independent runs, with the standard deviation in parentheses. Rows represent the evaluator agent's value, and columns represent the evaluated agent's value.}
    \label{fig:japanese_hobby_matrix}
\end{figure*}

The mutual evaluation scores under all conditions in this experiment are shown in Figures \ref{fig:english_house_matrix} and \ref{fig:english_hobby_matrix} for English and Figures \ref{fig:japanese_house_matrix} and \ref{fig:japanese_hobby_matrix} for Japanese. 
For each condition, we conducted 10 independent dialogue runs and report the mean and standard deviations of mutual evaluation score. 
As indicated by the row and column labels in Figures \ref{fig:english_house_matrix}–\ref{fig:japanese_hobby_matrix}, the matrices also indicate which agent evaluates which.
In the diagonal cells, the evaluator and the evaluated share the same value; therefore, the cell means are based on twice as many data points as the off-diagonal cells.
These results indicate a tendency toward higher trust and interpersonal closeness scores between agents with greater value similarity. 
For instance, in Figure \ref{fig:english_house_trust_basic}, the agent pairings that received the highest trust score of 5.0 were those with highly similar values, such as universalism–universalism and tradition evaluating conformity. 
Conversely, one of the pairings with the lowest score of 1.0 (Figure \ref{fig:english_house_trust_basic}) was hedonism evaluating conformity; these two values are in opposing positions within Schwartz's theory of basic values.

We categorized the similarity of values between agents into three levels (high, medium, and low) based on Schwartz's theory of basic values and calculated the average mutual evaluation scores for each level to quantitatively evaluate these visual observations (Tables \ref{tab:value_similarity_trust} and \ref{tab:value_similarity_ios}). 
To test whether evaluation scores increased monotonically with the degree of value similarity, Jonckheere–Terpstra tests were conducted separately for each language (English/Japanese), dialogue task (Housing/Hobbies), and value classification (basic values/higher-order values). 
Across all patterns, the results revealed significant positive trends, indicating that greater value similarity was associated with higher ratings of both trust and IOS (all p < .001; see Tables \ref{tab:value_similarity_trust_JonckTest} and \ref{tab:value_similarity_ios_JonckTest} for details).

We conducted two-way ANOVAs with language and topic as factors for each evaluation metric (trust / IOS) and each value classification (basic / higher-order) to statistically examine the effects of language and topic on evaluation scores.
Across all analyses, a consistent main effect of topic was observed: dialogues on the topic of ``housing'' yielded higher trust and IOS scores than those on ``hobbies'' (Trust–basic: $F(1, 4396) = 139.21$, $p < .001$; Trust–higher-order: $F(1, 796) = 29.15$, $p < .001$; IOS–basic: $F(1, 4396) = 446.60$, $p < .001$; IOS–higher-order: $F(1, 796) = 41.99$, $p < .001$).
Regarding the effect of language, significantly higher scores were obtained in Japanese than in English for Trust–basic, IOS–basic, and IOS–higher-order (Trust–basic: $F(1, 4396) = 396.38$, $p < .001$; IOS–basic: $F(1, 4396) = 678.73$, $p < .001$; IOS–higher-order: $F(1, 796) = 12.51$, $p < .001$). In addition, the topic $\times$ language interaction was significant for Trust–basic ($F(1, 4396) = 23.20$, $p < .001$), IOS–basic ($F(1, 4396) = 264.23$, $p < .001$), and IOS–higher-order ($F(1, 796) = 5.25$, $p = .022$). By contrast, for Trust–higher-order, neither the main effect of language ($F(1, 796) = 0.01$, $p = .938$) nor the interaction ($F(1, 796) = 0.001$, $p = .974$) reached significance.

Furthermore, we performed a correlation analysis to investigate the relationship between the evaluation patterns at basic and higher-order values (Table \ref{tab:corr_basic_higher}). 
The results demonstrated a statistically significant positive correlation in most cases. 
However, no significant correlation was found for interpersonal closeness in collaborative decision-making about the housing task in Japanese ($r=.2960$, $p=.2657$).

\begin{table}[ht]
    \centering
    \resizebox{\textwidth}{!}{%
    \begin{tabular}{@{}ll cc cc cc cc@{}}
    \toprule
    & & \multicolumn{4}{c}{\textbf{English}} & \multicolumn{4}{c}{\textbf{Japanese}} \\
    \cmidrule(lr){3-6} \cmidrule(lr){7-10}
    & & \multicolumn{2}{c}{Housing} & \multicolumn{2}{c}{Hobbies} & \multicolumn{2}{c}{Housing} & \multicolumn{2}{c}{Hobbies} \\
    \cmidrule(lr){3-4} \cmidrule(lr){5-6} \cmidrule(lr){7-8} \cmidrule(lr){9-10}
    \textbf{Value Similarity} & & Basic & \datacellt{Higher}{-order} & Basic & \datacellt{Higher}{-order} & Basic & \datacellt{Higher}{-order} & Basic & \datacellt{Higher}{-order} \\
    \midrule
    \multirow{2}{*}{High} & identical & 4.517 & 4.726 & 3.702 & 4.059 & 4.152 & 4.204 & 3.869 & 3.741 \\
    & same dimension & 3.969 & --- & 3.152 & --- & 3.969 & --- & 3.741 & --- \\
    Medium & & 2.898 & 3.158 & 2.267 & 2.817 & 3.620 & 3.551 & 3.228 & 3.038 \\
    Low & & 2.063 & 2.507 & 1.699 & 2.116 & 3.426 & 2.743 & 3.390 & 2.259 \\
    \bottomrule
    \end{tabular}
    }
    \caption{Relationship between value similarity and mutual evaluation results (Trust). Means are reported by similarity level.}
    \label{tab:value_similarity_trust}
\end{table}

\begin{table}[ht]
    \centering
    \resizebox{\textwidth}{!}{%
    \begin{tabular}{@{}ll cc cc cc cc@{}}
    \toprule
    & & \multicolumn{4}{c}{\textbf{English}} & \multicolumn{4}{c}{\textbf{Japanese}} \\
    \cmidrule(lr){3-6} \cmidrule(lr){7-10}
    & & \multicolumn{2}{c}{Housing} & \multicolumn{2}{c}{Hobbies} & \multicolumn{2}{c}{Housing} & \multicolumn{2}{c}{Hobbies} \\
    \cmidrule(lr){3-4} \cmidrule(lr){5-6} \cmidrule(lr){7-8} \cmidrule(lr){9-10}
    \textbf{Jonckheere--Terpstra test results} & & Basic & \datacellt{Higher}{-order} & Basic & \datacellt{Higher}{-order} & Basic & \datacellt{Higher}{-order} & Basic & \datacellt{Higher}{-order} \\
    \midrule
    JT  &                 & 312999 & 10577 & 1199899 & 39666 & 2619927 & 89071 & 4564026 & 157416 \\
    $z$ &                 & 18.085 & 9.773 & 22.136 & 11.593 & 24.230 & 14.044 & 25.866 & 15.852 \\
    $p$ &                 & $<.001$ & $<.001$ & $<.001$ & $<.001$ & $<.001$ & $<.001$ & $<.001$ & $<.001$ \\
    \bottomrule
    \end{tabular}
    }
    \caption{Jonckheere--Terpstra test results (Trust). }
    \label{tab:value_similarity_trust_JonckTest}
\end{table}

\begin{table}[ht]
    \centering
    \resizebox{\textwidth}{!}{%
    \begin{tabular}{@{}ll cc cc cc cc@{}}
    \toprule
    & & \multicolumn{4}{c}{\textbf{English}} & \multicolumn{4}{c}{\textbf{Japanese}} \\
    \cmidrule(lr){3-6} \cmidrule(lr){7-10}
    & & \multicolumn{2}{c}{Housing} & \multicolumn{2}{c}{Hobbies} & \multicolumn{2}{c}{Housing} & \multicolumn{2}{c}{Hobbies} \\
    \cmidrule(lr){3-4} \cmidrule(lr){5-6} \cmidrule(lr){7-8} \cmidrule(lr){9-10}
    \textbf{Value Similarity} & & Basic & \datacellt{Higher}{-order} & Basic & \datacellt{Higher}{-order} & Basic & \datacellt{Higher}{-order} & Basic & \datacellt{Higher}{-order} \\
    \midrule
    \multirow{2}{*}{High} & identical & 6.340 & 6.175 & 5.850 & 6.150 & 6.010 & 5.938 & 5.925 & 5.900 \\
    & same dimension & 5.875 & --- & 4.088 & --- & 5.912 & --- & 5.819 & --- \\
    Medium & & 5.194 & 5.438 & 3.612 & 4.550 & 5.737 & 5.925 & 5.435 & 5.400 \\
    Low & & 4.781 & 5.400 & 2.308 & 3.350 & 5.808 & 5.450 & 5.592 & 4.725 \\
    \bottomrule
    \end{tabular}
    }
    \caption{Relationship between value similarity and mutual evaluation results (IOS). Means are reported by similarity level.}
    \label{tab:value_similarity_ios}
\end{table}

\begin{table}[ht]
    \centering
    \resizebox{\textwidth}{!}{%
    \begin{tabular}{@{}ll cc cc cc cc@{}}
    \toprule
    & & \multicolumn{4}{c}{\textbf{English}} & \multicolumn{4}{c}{\textbf{Japanese}} \\
    \cmidrule(lr){3-6} \cmidrule(lr){7-10}
    & & \multicolumn{2}{c}{Housing} & \multicolumn{2}{c}{Hobbies} & \multicolumn{2}{c}{Housing} & \multicolumn{2}{c}{Hobbies} \\
    \cmidrule(lr){3-4} \cmidrule(lr){5-6} \cmidrule(lr){7-8} \cmidrule(lr){9-10}
    \textbf{Jonckheere--Terpstra test results} & & Basic & \datacellt{Higher}{-order} & Basic & \datacellt{Higher}{-order} & Basic & \datacellt{Higher}{-order} & Basic & \datacellt{Higher}{-order} \\
    \midrule
    JT  &                 & 314945 & 9321 & 1226302 & 39564 & 2566861 & 82703 & 4422563 & 145235 \\
    $z$ &                 & 19.324 & 8.000 & 24.475 & 12.657 & 23.919 & 13.076 & 24.995 & 14.636 \\
    $p$ &                 & $<.001$ & $<.001$ & $<.001$ & $<.001$ & $<.001$ & $<.001$ & $<.001$ & $<.001$ \\
    \bottomrule
    \end{tabular}
    }
    \caption{Jonckheere--Terpstra test results (IOS).}
    \label{tab:value_similarity_ios_JonckTest}
\end{table}

\begin{table}[ht]
    \centering
    \begin{tabular}{@{}l ll ll@{}}
    \toprule
    & \multicolumn{2}{c}{\textbf{English}} & \multicolumn{2}{c}{\textbf{Japanese}} \\
    \cmidrule(lr){2-3} \cmidrule(lr){4-5}
    & Housing & Hobbies & Housing & Hobbies \\
    \midrule
    trust & 0.9444** & 0.9002** & 0.8343** & 0.7925** \\
    IOS   & 0.8021** & 0.8099** & 0.2960 & 0.5180* \\
    \bottomrule
    \end{tabular}
    \caption{Pearson's correlation coefficient between basic values and higher-order values. ($^{*}p < .05$, $^{**}p < .01$)}
    \label{tab:corr_basic_higher}
\end{table}

\clearpage

\section*{Discussion}

The results of the preliminary experiment on value controllability revealed that the perspective controllability scores varied depending on the target language model, prompt format, and language (Tables \ref{tab:controllability_basic} and \ref{tab:controllability_higher}). 
First, a comparison of the performance of the models revealed that Gemini achieved the highest average scores in both English and Japanese. 
This suggests that Gemini has a better ability to generate responses aligned with a specified perspective than the other models under the conditions of this study.
Next, we observed two trends regarding the influence of the prompt format. 
First, controlling the models based on the ten basic values was consistently more challenging than controlling the models based on the four higher-order values. 
It is conceivable that, because the latter represents broader and more abstract concepts, the models can more easily grasp them as general directives, leading to improved controllability. 
Second, the effects of grammatical persons varied among the models. 
For Gemini, second-person prompts yielded greater controllability than third-person prompts. 
By contrast, GPT exhibited the opposite trend, in which third-person prompts tended to result in higher controllability. 
Furthermore, comparing across languages, we also found that English exhibits higher controllability than Japanese, as indicated by the results of the t-test.
This is likely attributable to the fact that the training data for several large language models are predominantly in English. 
Consequently, we infer that the models' comprehension and adherence to instructions were higher for English than for Japanese. 
The existence of this linguistic bias poses a significant challenge for applying LLMs to value-based systems in non-Anglophone contexts.

A further comparison between the English and Japanese results revealed a notable difference: whereas scores were stably calculated in the English experiments, the Japanese experiments exhibited a significant number of missing values, which corresponded to response refusals by the models (Tables \ref{tab:controllability_basic} and \ref{tab:controllability_higher}). 
Communication in Japan is often discussed as a classic example of a high-context culture \cite{nagaishi2022exploring}, and recent studies have indicated that LLMs reflect the social biases present in training data \cite{sheng2021societal}. 
Based on these findings, it is conceivable that LLMs learn not only vocabulary and grammatical structures but also statistically internalize communication styles rooted in a high-context culture, such as avoiding direct assertions and circumventing conflict, during their training on Japanese text data. 
Therefore, we infer that achieving effective value control in the Japanese context requires sophisticated prompt engineering that goes beyond merely presenting values. 
Such prompts must also signal to the model the meta-level situational judgment such that ``an explicit and assertive evaluation is necessary in this context.'' 
Thus, our findings suggest that the value control of LLMs is a localization challenge that necessitates a deep understanding of the linguistic and cultural context in which the model operates.

Subsequently, the results of the main experiment are discussed. 
As summarized in Tables \ref{tab:value_similarity_trust}, \ref{tab:value_similarity_trust_JonckTest}, \ref{tab:value_similarity_ios}, and  \ref{tab:value_similarity_ios_JonckTest}, a higher similarity in values, based on Schwartz's theory of basic values, was associated with higher mutual trust ratings and a greater sense of interpersonal closeness. 
This finding is consistent with a broad range of previous studies demonstrating the importance of value similarity in trust and relationship-building \cite{earle1995social,siegrist2000salient,murstein1970stimulus}. 
Therefore, our results can be interpreted as successful replication of these findings within the computational framework of mutual evaluation. 
Furthermore, as summarized in Table \ref{tab:corr_basic_higher}, a positive correlation was observed between the evaluation patterns of the ten basic values and the four higher-order values under most conditions. 
This finding suggests that Schwartz's theory \cite{schwartz1992universals,schwartz2012overview} was also reproduced within the mutual evaluation framework employed in this study.

Furthermore, Figures \ref{fig:english_house_matrix}, \ref{fig:english_hobby_matrix}, \ref{fig:japanese_house_matrix}, and \ref{fig:japanese_hobby_matrix} illustrate the influence of each value on mutual evaluation. 
Specifically, a tendency was observed for mutual trust evaluations to be higher among agents who prioritize values classified under ``self-transcendence'' and ``conservation'' in Schwartz's model (e.g., universalism, benevolence, tradition, conformity, and security), as shown, for instance, in Figure \ref{fig:english_hobby_trust_basic}. 
This suggests that agents holding these values, who tend to prioritize harmony with others and social norms over personal gain, contributed to establishing trusting relationships. 
By contrast, the results indicated that agents with values oriented toward personal interest and achievement, such as ``self-enhancement'' and ``openness to change,'' had difficulty establishing trusting relationships, particularly with those holding dissimilar values. 
Although it has been highlighted in psychology that individuals who prioritize self-transcendence tend to adopt a cooperative stance and consequently build high-quality social relationships \cite{bojanowska2021individual}, this simulation revealed a distinct pattern in which specific values exert a strong influence, particularly on the formation of trust. 
Validating the findings of this study through future empirical research with human participants will provide new insights into the role of values in social sciences.

Differences in the mutual evaluations based on the dialogue topic were also observed. 
Two-way ANOVAs with language and topic as factors for each evaluation metric (trust / IOS) and each value classification (basic / higher-order) revealed that, across all patterns, the main effect of topic was significant. 
Specifically, in collaborative decision-making tasks about housing, evaluations of trust and IOS were significantly higher compared to dialogues about hobbies.
This finding can be explained by the relationship development process. 
Interpersonal relationships deepen from the initial stage of simple self-disclosure to a more advanced stage of sharing fundamental values, and the depth of the relationship is linked to trust \cite{altman1973social,rempel1985trust}. 
In the present study, dialogue about hobbies was assumed to mimic the superficial level of interaction characteristic of the initial stages of a relationship, whereas the cooperative decision-making task was considered to represent interaction at a deeper level, involving the sharing of values. 
Therefore, the difference in evaluation scores observed in this experiment suggests that topic of dialogue can simulate the process of relationship development.

Furthermore, differences in the mutual evaluations based on language were observed. 
Except for the evaluation of trust under higher-order values control, the ANOVA results demonstrated a significant main effect of language across all other patterns. 
Specifically, evaluations of trust and IOS were significantly higher in the Japanese context compared to the English context.
This trend suggests that the simulation reflects the tendency of the Japanese culture to emphasize harmony and relationships with others \cite{markus2014culture}. 
Moreover, the correlation analysis summarized in Table \ref{tab:corr_basic_higher} reveals lower correlation coefficients for Japanese than for English. 
Specifically, no significant correlation was found between the evaluation patterns of basic and higher-order values in cooperative decision-making regarding housing tasks. 
This may indicate that the structure of values depends on cultural context \cite{schwartz1992universals,markus2014culture}. 
Thus, the deviation in the pattern observed under Japanese conditions in this study may reflect this cultural dependency.

Interestingly, in the higher-order values control, the effect of language disappeared, and only the main effect of topic was observed for trust evaluation. 
This pattern may be explained by the difference in the level of abstraction between basic and higher-order values in Schwartz's theory. 
The ten basic values are relatively fine-grained and may carry culture-specific nuances, making trust judgments more sensitive to cultural signals conveyed through language. 
By contrast, the four higher-order values represent broader and more abstract motivational dimensions that are more universally shared across cultures~\cite{schwartz1992universals,schwartz2012refining}. 
When agents are guided by higher-order values, the influence of language as a cultural signal may attenuate, and trust judgments may instead rely more heavily on the concrete topic of dialogue. 
In contrast, IOS scores showed consistent language effects across both basic and higher-order values control. 
This divergence indicates that trust and IOS capture partially different aspects of interpersonal evaluation. 
While trust is the willingness to be vulnerable to another party based on positive expectations~\cite{mayer1995integrative}, the IOS scale is a measure of closeness represented by the perceived overlap between the self and another~\cite{aron1992inclusion}. 
The latter appears to remain sensitive to linguistic context even when values are framed at a more abstract level.

Finally, we address the limitations of this study and outline directions for future research. 
While the present study employed specific dialogue tasks to simulate relationship development, future work could involve experiments under more diverse or controlled conditions to disentangle the effects of the task setting itself.
Another compelling direction is to focus on the behavioral consistency and traits of individual agents. 
Although our work has centered on the influence of value similarity, investigating the correlation between an agent's tendency to be a lenient or strict evaluator and the ratings it receives from others presents a valuable avenue for future research.
Furthermore, a key future task will involve constructing a model that considers the interactions among multiple values. 
In the present experiment, the value provided to the LLM in the prompt was singular; therefore, the influence of other coexisting values could not be considered. 
However, humans often hold multiple values and make decisions based on trade-offs between them \cite{tetlock1986value}. Therefore, modeling this value plurality is expected to enable a more realistic value simulation.

We propose the development of a more comprehensive agent model that integrates both stable base values and more flexible ``prioritized'' values that may shift depending on emotions and contextual reactions.
Human decision-making is influenced not only by values but also by various cognitive and affective factors, such as momentary emotions and the inference of the intentions of others \cite{gratch2004domain}. 
Incorporating these elements into the agent architecture is likely to enable more human-like simulations under realistic conditions.

In interpreting our findings, it is important to consider the choice of value framework. 
While the refined theory distinguishes 19 more specific values, these can be recombined into the original ten basic values \cite{schwartz2012refining}. 
Many studies still employ measures based on the ten values (e.g., \cite{kovavc2023large,kovavc2024stick}), and we therefore adopted this framework to ensure comparability with previous research. 
Nevertheless, we acknowledge that the more fine-grained value model represents a promising direction for future extensions.

This study was confined to two languages, English and Japanese. 
This choice was theoretically motivated, as it allowed for a focused comparison between a dominant training language of LLMs (English) and a high-context language with relatively less representation in training data (Japanese). 
A significant direction for future research will be to expand the experimental framework to a broader range of languages and cultural contexts, thereby testing the universality of the value similarity principle in agent simulations across diverse cultural settings.

Another important limitation concerns the motivational asymmetry between humans and LLM agents. 
Humans are strongly motivated to evaluate trust cautiously, since misplaced trust can threaten survival, well-being, and social relationships \cite{bell2019trust,thoresen2018loss}. 
In contrast, LLMs lack such intrinsic motivations. 
Recent research, however, suggests promising directions for bridging this gap. 
Studies on agent architectures that integrate internal states and reflection \cite{park2023generative,shinn2023reflexion}, research on intrinsic motivation in artificial intelligence \cite{oudeyer2007intrinsic,schmidhuber2010formal}, and the use of explicit prompting strategies point toward ways of endowing agents with human-like motivational structures. 
Such models would enable simulations that more closely align with the human tendency to be cautious when trust carries real stakes, thereby enhancing the realism and explanatory power of LLM-based trust research.

In addition, we asked the LLM agents to make evaluations using the IOS diagrams. 
Previous studies have reported that multimodal LLMs can analyze images and extract their psychological meaning (e.g., \cite{zhang2024psydraw,wu2025vs}). 
Moreover, the gemini-1.5-flash model used in our main experiment has been described in Google's technical documentation as having strong visual processing capabilities \cite{team2024gemini}. 
At the same time, unlike human participants who may attribute rich social meaning to the IOS diagrams, the LLM's judgments are likely constrained by its visual–textual alignment. 
We therefore note this as a limitation of the present study.

\section*{Methods}

\subsection*{Overview}

This study comprises two parts: a preliminary experiment to evaluate the value controllability of LLMs and a main experiment that performs dialogue simulations between LLM agents endowed with specific values. 
The preliminary experiment assessed whether LLMs could express specific values assigned via prompts (i.e., value controllability). 
This step was crucial for ensuring the validity of the main experiment, which used agents with diverse values. 
The proposed methodology builds upon the work of Kovac et al. \cite{kovavc2023large} and extends it to three key areas: adaptation to the Japanese context, addition of value definitions to prompts, and incorporation of basic human values from Schwartz's theory. 
In the main experiment, we generated agent pairs holding specific values using the LLM whose value controllability was confirmed in the preliminary experiment. 
After these pairs engaged in dialogue, we used a questionnaire to measure their mutual trust and interpersonal closeness to analyze the impact of value similarity on relationship formation.

\subsection*{Preliminary Experiment: Evaluation of Value Controllability of LLMs}

This preliminary experiment aims to quantitatively evaluate the value controllability of LLMs through prompting. 
We evaluate the outputs of LLMs using the ``perspective controllability'' metric to measure their ability to alter their characteristics based on context \cite{kovavc2023large}. 
This study also investigated linguistic differences by comparing performance in English and Japanese environments. 
This section details the prompts used in the experiment, methodology for measuring value controllability, and LLMs subjected to evaluation.

\subsubsection*{prompt}

In this experiment, we systematically evaluate the influence of multiple prompt elements on value controllability. 
The following factors were examined: (1) the person of the prompt (second-/third-person), (2) classification of the target value, (3) presence or absence of a definition for the value, and (4) prompt type (system/user).

Base prompts were formulated from both the second- and third-person perspectives. 
The templates used were as follows:

\begin{quote}
\textbf{Second-person prompt:}
You are a person attributing extremely more importance to \{value\}.\\
\textbf{Third-person prompt:}
The following are answers from a person attributing extremely more importance to high \{value\}.
\end{quote}

Here, \{value\} is substituted with value labels based on Schwartz's theory of basic values \cite{schwartz1992universals, schwartz2012overview}. 
In this study, we evaluated the controllability of each of the ten basic values defined by the theory (e.g., power, achievement, and benevolence) and the four higher-order values that encompass them (e.g., self-enhancement and self-transcendence). 
When controlling for one of the ten values, a single basic value was substituted into \{value\}. 
When controlling for one of the four higher-order values, the constituent basic values under that higher-order concept were enumerated and substituted into \{value\}.

Furthermore, we establish a condition in which a definition of value is added to the prompt based on proposals from related studies to reduce the ambiguity of abstract concepts, such as values \cite{ciupinska2024awareprompt}. 
The values were defined based on previous studies \cite{sagiv1995value}. 
For instance, when controlling for power, the second-person prompt with a definition is as follows: ``You are a person attributing extremely more importance to power (social status and prestige, control or dominance over people and resources (social power, authority, wealth)).'' 
In addition, when controlling for self-enhancement, the third-person prompt without definitions is as follows: ``The following are answers from a person attributing extremely more importance to power, achievement.''
Additionally, we compared two methods of prompt type: providing the prompt as a system prompt and including it at the beginning of the user prompt.

\subsubsection*{Measurement Method}

We prompted the LLM to respond to a psychological scale designed to measure values and scored the results to evaluate value controllability. 
We used the portrait values questionnaire (PVQ), which is based on Schwartz's theory of basic values \cite{schwartz2001extending, schwartz2021repository}. 
The PVQ comprises 40 items, each of which describes a portrait of a person with specific values. 
Respondents rated the similarity of each portrait to themselves on a 6-point scale.

For the experimental procedure, the 40 portrait items were independently presented to the LLMs in random order to generate responses. 
This process was repeated 50 times to ensure robustness. 
Each API call to the LLM is independent, and the conversation history variables are reset each time, so that no systematic drift could occur within the same foundational LLM.
Considering that extracting subjective information from LLMs can be challenging \cite{kobayashi2024extraction}, if valid responses were not obtained for 10\% or more of the items in the first iteration, the trial for that model and prompt combination was aborted, and the score was not calculated.

The perspective controllability score was calculated based on the obtained responses. 
The perspective controllability score $C^M_V$ for a language model $M$ on a value $V$ is defined as follows:

\begin{equation}
   C_V^M = \underset{d \in V}{\text{mean}(s_d)} - \underset{d' \notin V}{\text{mean}(s_{d'})}
   \label{eq:controllability}
\end{equation}

Here, $s_d$ denotes the response score for portrait $d$, normalized to a range of [0, 1] and $C_V^M$ represents the difference obtained by subtracting the mean response score for the set of portraits unrelated to the target value $V$ from that for the set of portraits related to $V$. 
A higher score indicates greater value controllability.
In this study, to evaluate both the ten basic values and the four higher-order values of Schwartz's theory, we calculate $C^M_{l}$ (the average of $C_V^M$ across the ten basic values) and $C^M_{h}$ (the average of $C_V^M$ across the four higher-order values) as the overall value controllability measures for model $M$.

\subsubsection*{LLMs}

We evaluated the following three LLMs to compare the performance differences across models: gpt-4o, gemini-1.5-flash, and claude-3.5-sonnet. 
The generation parameters for each model were maintained at their default settings, except for the maximum output length (max\_tokens), which was set to 1000.

\subsection*{Main Experiment: Effect of Value Similarity on Agent Interaction and Mutual Evaluation}

In the main experiment, we investigated the influence of value similarity on the mutual evaluation of LLM agents following dialogue. 
We generated agents using the model and prompt combination that demonstrated the highest value controllability in the preliminary experiment, and simulated dialogues between agents assigned specific values. 
After the dialogue, the agents evaluated each other, and we analyzed the results in relation to their value similarity. 
This section describes the interaction design, metrics used for the mutual evaluation, and data analysis methods.

\subsubsection*{Interaction Design}

We prepared two LLM agents, each assigned a specific value by prompting and simulating dialogue between them. 
Value combinations were comprehensively generated based on Schwartz's theory of basic values, covering ten basic values and four higher-order values. 
Specifically, for the basic values, we configured 55 pairs in total, comprising 10 pairs with identical values and 45 pairs with different values ($_{10}C_2 = 45$).
Similarly, for the higher-order values, we configured 10 pairs in total (four pairs with identical values and $_{4}C_2 = 6$ pairs with different values).

Furthermore, we established the following two tasks as topics for dialogue, which simulate the initial stage of the relationship-building process and a later stage after the relationship has deepened, respectively.

\begin{enumerate}
    \item \textbf{Dialogue about hobbies:} The agents exchange information about their respective hobbies. According to the social penetration theory, self-disclosure in the initial stages of relationship building is characterized by a low breadth and depth \cite{altman1973social}. We selected hobbies as a topic for casual dialogue, where values can be easily expressed because hobbies are not only a common topic for initial self-disclosure but can also contain information reflecting the values of an individual \cite{sakamoto2025effectiveness}.
    \item \textbf{Collaborative decision-making about housing:} The two agents discuss a scenario in which they must jointly select a house to live in. This task simulates a more advanced stage of relationship development in which value similarity plays a crucial role \cite{murstein1970stimulus}. Based on prior decision-making studies, we presented price, location, area in square meters, number of rooms, and the condition of the unit as criteria for selecting a residence, prompting a discussion on the factors that should be prioritized \cite{korhonen1990choice}.
\end{enumerate}

Dialogue was initiated with an opening statement from one of the agents. 
For topic 1 (hobbies), the initial utterance was, ``Let's talk about each other's hobbies. What are your hobbies?'' 
For topic 2 (housing), the initial utterance was, ``Let's discuss which of the following we should prioritize for the house we're going to live in together: price, location, area in square meters, number of rooms, and the condition of the unit.'' 
Subsequently, the agents took turns speaking, and the dialogue was concluded once ten turns (five from each agent) were completed.

In each dialogue, the two agents were implemented as independent calls to LLM. 
For every turn, system prompt was set separately, and no session-level hidden states or memory were shared across agents. 
Instead, the dialogue history was explicitly reconstructed at each step and inserted into the prompt for the relevant agent. 
This design ensured that each agent's utterance was generated independently, while still maintaining conversational coherence through the prompt.
Furthermore, no memory or hidden states were shared across the two dialogue tasks.

\subsubsection*{Mutual Evaluation Metrics}

After the dialogue was concluded, each agent was presented with a conversation history and tasked with evaluating its counterpart. 
Trust and interpersonal closeness were used as evaluation metrics:

\begin{itemize}
    \item \textbf{Trust:} Value similarity influences trust \cite{siegrist2000salient}. We used a three-item scale from a study on value similarity and trust to measure trust, asking the agents to rate their counterpart on a 5-point scale for each item: “Can be trusted,” “Can be relied upon,” and “Can be entrusted” \cite{nakayachi2014method, Yokoi2018effects}.
    \item \textbf{Interpersonal closeness:} We used the IOS scale to measure the closeness of the relationship with a counterpart \cite{aron1992inclusion}. The IOS scale is a measure in which the agent selects one of seven diagrams, each depicting two circles with varying degrees of overlap, that best represents its relationship with the other agent.
\end{itemize}

This process of interaction and mutual evaluation was performed independently ten times for each pair of values and each dialogue task to ensure the robustness of the results.

\subsubsection*{Data Analysis Methods}

The mutual evaluation data were analyzed from the following two perspectives:

\begin{enumerate}
    \item \textbf{Relationship between Value Similarity and Mutual Evaluation:} 
    It is known that Schwartz's values are not independent but rather possess a circular structure of motivational relationships. Therefore, in this study, we did not treat each value as independent and instead defined the value similarity between agents based on this circular model \cite{schwartz1992universals, schwartz2012overview}. In this model, values close to the circular continuum are motivationally compatible, whereas distant values are in conflict. Based on this principle, we classified value similarity into the following three levels. We analyzed the impact of value similarity on evaluations by comparing mutual evaluation scores (trust and interpersonal closeness) across the three similarity levels.

    \begin{itemize}
        \item \textbf{High Similarity:} Pairs with identical values (e.g., power–power) and pairs whose values belong to the same higher-order dimension in the circular model (e.g., power–achievement). In the context of higher-order values, this level comprises only identical pairs.
        \item \textbf{Low Similarity:} Pairs with values belonging to opposing higher-order dimensions in the circular model (e.g., power–universalism).
        \item \textbf{Medium Similarity:} All other combinations not classified as high or low.
    \end{itemize}
    
    \item \textbf{Correlation Analysis between Basic Value and Higher-Order Value Results:} 
    We verified whether controlling for a higher-order value reflects the collective tendencies of its corresponding basic values. 
    To achieve this, we first created an aggregated higher-order value evaluation matrix by averaging the evaluation scores of the basic value pairs belonging to each higher-order dimension using the results from the basic value control experiments. 
    For example, the aggregated score for the ``self-enhancement to self-transcendence'' dimension was calculated by averaging the mutual evaluation scores from the four corresponding basic value pairs: power to universalism, power to benevolence, achievement to universalism, and achievement to benevolence.
    Next, we calculated Pearson's correlation coefficient between this aggregated matrix and the direct higher-order value evaluation matrix obtained by directly controlling for the higher-order values. 
    Each evaluation matrix was flattened into a one-dimensional vector to calculate the correlation.
\end{enumerate}

\section*{Data Availability}
The experimental code, prompts, and raw data analyzed in this study are available from the following repository: \url{https://osf.io/7jn8x/?view_only=a1afa8564ac04a4481ccb03b3f0dd555}

\bibliography{sample}

\section*{Funding Declaration}

This work was supported by JSPS KAKENHI Grant Number 22K17949, JST PRESTO Grant Number JPMJPR23I2, and JST BOOST, Japan Grant Number JPMJBS2402.

\section*{Author contributions statement}

All authors (Y.S., T.U., H.I.) contributed to the study's conception and design.
Y.S. collected data, and Y.S. and T.U. analyzed the results.
Y.S. and T.U. prepare the material, and all authors approved the final manuscript.

\section*{Competing interests}
The authors declare no competing interests.

\end{document}